%% file: main.tex
\documentclass{bmvc2k}

\usepackage{graphicx}
\usepackage{color, xcolor}
\usepackage{multicol, multirow}
\usepackage{microtype}
\usepackage{adjustbox}
\usepackage{booktabs}
\usepackage{bm}
\usepackage{algorithm, algpseudocode}
\usepackage{amsmath, amsfonts, amssymb}
\usepackage{overpic}
\usepackage{enumitem}
\usepackage{contour}
\contourlength{0.5pt}
\usepackage{pgf,tikz}
\usetikzlibrary{shapes.geometric}
\usepackage{pgfplots}
\pgfplotsset{compat=1.7}

\title{Detect, Augment, Compose, and Adapt: Four Steps for Unsupervised Domain Adaptation in Object Detection}

\addauthor{Mohamed L. Mekhalfi}{mmekhalfi@fbk.eu}{1}
\addauthor{Davide Boscaini}{dboscaini@fbk.eu}{1}
\addauthor{Fabio Poiesi}{poiesi@fbk.eu}{1}

\addinstitution{Technologies of Vision Lab \\Fondazione Bruno Kessler, Trento, Italy}

\runninghead{Mekhalfi, Boscaini, Poiesi}{DACA}

\def\etal{\emph{et al}\bmvaOneDot}
\def\ourmethod{DACA\xspace}

\begin{document}

\maketitle

\input{sections/abstract}
\input{sections/intro}
\input{sections/related}

\input{sections/method}
\input{sections/results}
\input{sections/conclusions}
\input{sections/acknow}

\bibliography{biblio}

\input{sections/supp}

\end{document}

%% file: sections/abstract.tex
\begin{abstract}
Unsupervised domain adaptation (UDA) plays a crucial role in object detection when adapting a source-trained detector to a target domain without annotated data. In this paper, we propose a novel and effective four-step UDA approach that leverages self-supervision and trains source and target data concurrently. We harness self-supervised learning to mitigate the lack of ground truth in the target domain. 
Our method consists of the following steps: 
(1) identify the region with the highest-confidence set of detections in each target image, which serve as our pseudo-labels;
(2) crop the identified region and generate a collection of its augmented versions;
(3) combine these latter into a composite image;
(4) adapt the network to the target domain using the composed image.
Through extensive experiments under cross-camera, cross-weather, and synthetic-to-real scenarios, our approach achieves state-of-the-art performance, improving upon the nearest competitor by more than 2\% in terms of mean Average Precision (mAP).
The code is available at \url{https://github.com/MohamedTEV/DACA}. 
\end{abstract}

%% file: sections/intro.tex
\section{Introduction}\label{sec:intro}

Domain Adaptation in object detection aims to adapt an object detection model, trained on a data distribution (source), to perform well on a different data distribution (target domain). 
The challenge of this process arises not only from domain shift but also from factors such as the high variability in object statistics within the target domain. 
For example, target objects with a different appearance from that of source objects are more difficult to adapt.
Another challenge is the data bias across domains, such as the disparity in the quantity of source objects versus the abundance of objects in the target domain.
When the target domain possesses a few ground-truth annotations, these can be used as initial seeds to enhance the model's generalization capabilities on the target, which has been shown to be effective in past research works~\cite{wang2019few, kang2019few}. 
However, in real applications the target domain often lacks annotations, and this leads to Unsupervised Domain Adaptation (UDA) techniques~\cite{SCUDA}.

\input{images/teaser/teaser3}

UDA can be tackled through style transfer techniques to mitigate domain shift~\cite{curriculum, rodriguez2019domain}. 
Style transfer involves modifying the appearance of a source image to resemble the aesthetic of a target image, while retaining the semantic content.
Then, a detector is trained on this intermediate data domain and deployed on the target domain. 
Despite its simplicity, this strategy can encounter limitations when the target domain's style is inconsistent.
In particular, the source may successfully emulate the style of a subset of the target while remaining misaligned with other underlying subsets.
An alternative approach that can address this issue is feature alignment, which tackles the problem at the feature level rather than at the image appearance level~\cite{align1, align2, mga, xiong2022source}.
In addition to feature alignment, other UDA methods can adapt a detector's model by using detections as pseudo-labels that are generated from target images~\cite{confmix, SCUDA}.
However, a too inclusive pseudo-label selection (i.e.~lenient selection criterion) may accumulate false positives, causing unreliable self-training.
Conversely, implementing strict measures for pseudo-label selection may lead to few true positives, making the detector to remain overly reliant on the source domain.
In either scenario, the pseudo-labels could mislead the detector or cause overfitting.
Recent research has shown that data augmentation can yield plausible improvements in classification and segmentation tasks~\cite{cutout, cutmix, mixup, wu2020dual}.
Approaches based on data-augmentation tackle the distribution shift problem at input level, without requiring modifications to the detector's architecture.
Our approach is a novel technique that belongs to the family of augmentation-based approaches.

In this paper, we propose a novel approach for UDA in object detection.
Our approach is based on four simple processing steps: \textit{detect}, \textit{augment}, \textit{compose}, and \textit{adapt}.
We refer to our approach as \ourmethod.
\ourmethod is inspired by the idea that adaptation can be made effective by generating high-quality pseudo-labels from the target and using them to supervise augmented versions of the target itself to align the model to the target data distribution.
Specifically, we exploit the region of target images with the most confident object detections to gather pseudo-labels for self-training. 
We create a composite image by applying a collection of random augmentations to the original target image region. 
This mechanism allows the detector to self-train with confident pseudo-labels applied to augmented images under domain shift.
We evaluate \ourmethod on the popular benchmarks Cityscapes~\cite{cityscapes}, FoggyCityscapes~\cite{foggycityscapes}, Sim10K~\cite{sim10k}, and KITTI~\cite{kitti}.
\ourmethod achieves state-of-the-art performance, improving upon the nearest competitor by more than 2\% in terms of mean Average Precision (mAP).
Fig.~\ref{fig:teaser} illustrates the idea of \ourmethod.
To summarize, our contributions are:
\begin{itemize}[noitemsep,topsep=0pt]
    \item Our approach is the first alternative to mix up approaches that does not mix images from different domains, but instead generates difficult and informative composite images only from the unsupervised target images.
    \item We devise a novel approach for UDA based on self-supervision. 
    \ourmethod generates the composite image based on augmented versions of the target image region with the most confident detections, making the adaptation more effective.
\end{itemize}

%% file: images/teaser/teaser3.tex
\begin{figure}[t!]
\centering
\begin{overpic}[width=\textwidth]{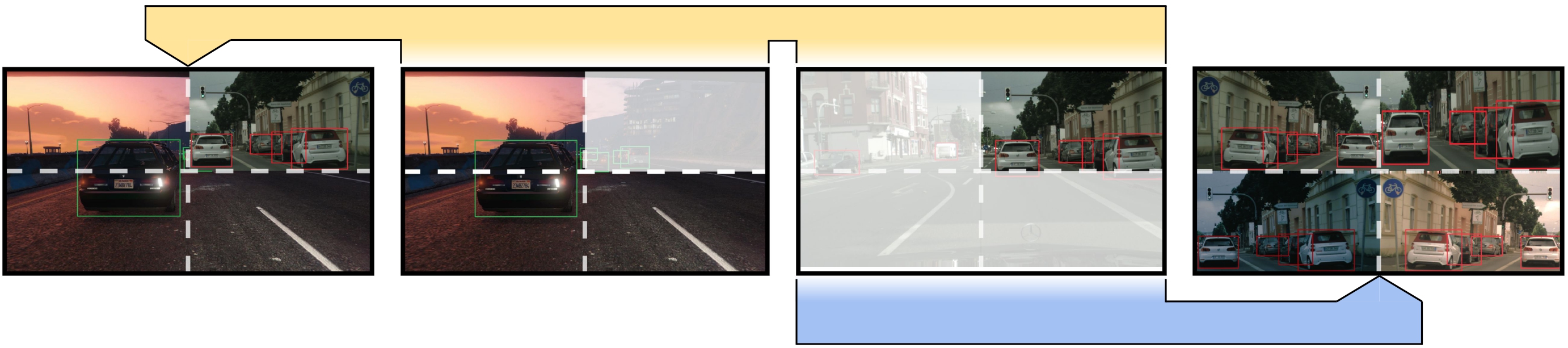}
    \put(6,-2.5){\small ConfMix~\cite{confmix}}
    \put(31,-2.5){\small Source image}
    \put(56,-2.5){\small Target image}
    \put(81,-2.5){\small \ourmethod (Ours)}
    \put(0.6,15.75){\small \color{black}{$S_{11}$}}
    \put(0.6,5.5){\small \color{white}{$S_{21}$}}
    \put(20.4,15.75){\small \color{white}{$T_{12}$}}
    \put(20.4,5.5){\small \color{white}{$S_{22}$}}
    \put(26.,15.75){\small \color{black}{$S_{11}$}}
    \put(26,5.5){\small \color{white}{$S_{21}$}}
    \put(45.5,15.75){\small \color{black}{$S_{12}$}}
    \put(45.5,5.5){\small \color{white}{$S_{22}$}}
    \put(51.25,15.75){\small \color{black}{$T_{11}$}}
    \put(51.25,5.5){\small \color{black}{$T_{21}$}}
    \put(71,15.75){\small \color{white}{$T_{12}$}}
    \put(71,5.5){\small \color{black}{$T_{22}$}}
    \put(76.5,15.75){\small \color{white}{$f_1(T_{12})$}}
    \put(76.5,5.5){\small \color{white}{$f_3(T_{12})$}}
    \put(92.1,15.75){\small \color{white}{$f_2(T_{12})$}}
    \put(92.1,5.5){\small \color{white}{$f_4(T_{12})$}}
    \put(26.5,19.5){\small Select, Crop, and Mix}
    \put(52.5,1.0){\small Select, Crop, Augment, and Compose}
\end{overpic}

\vspace{2mm} 
\caption{
Comparison between our method (\ourmethod) and ConfMix~\cite{confmix}. Green bounding boxes are ground-truth annotations, red bounding boxes are pseudo-detections. $S_{\text{row}\,\text{col}}$ are source regions, $T_{\text{row}\,\text{col}}$ are target regions. $T_{12}$ in this example corresponds to the most confident target region, while the remaining transparent regions of the target image are discarded. $f_1(T_{12})$, $f_2(T_{12})$, $f_3(T_{12})$, $f_4(T_{12})$ are augmented versions of $T_{12}$. 
Unlike \ourmethod, ConfMix requires source images to build the mixed image.
}

\label{fig:teaser}
\end{figure}
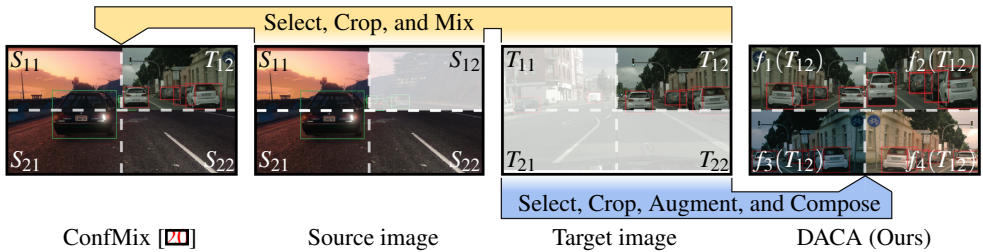

%% file: sections/related.tex
\section{Related Work}

\subsection{UDA for object detection}

Domain adaptation has been an active area of research to compensate for the distribution mismatch between a source and target data in classification and segmentation endeavours~\cite{wei2021metaalign, ghifary2016deep, zhang2019curriculum, kang2019contrastive, zhu2019multi, mahapatra2023unsupervised, wu2021dannet, kouw2019review}.
It has assumed several adaptation forms such as multi-source adaptation~\cite{multisource}, few-shot adaptation~\cite{fewshot}, and source-free adaptation~\cite{sourcefree}. 
For instance, Inoue \etal~\cite{inoue2018cross} tackle the problem of weakly-supervised domain adaptation where instance-level annotations are provided at the source and image-level annotations are envisioned at the target. 
Wang \etal~\cite{wang2023d2f2wod} leverage synthetic data to carry out the adaptation.

As per UDA in object detection, Zhu \etal~\cite{zhu2019adapting} align features of object-prone regions across disjoint domains.
Li \etal~\cite{li2022cross} present a student-teacher approach, where the student model ensures cross-domain consistent features via adversarial training.
SC-UDA~\cite{SCUDA} translates source images to an intermediate domain by CycleGAN to close the style gap with the target, while content gap is reduced by fusing multiple pseudo-detections produced with embedded stochastic inference.
Soviany \etal~\cite{curriculum} adopt curriculum learning, which aims to progressively adapt a source-trained model to a target, starting with easy images.
Image difficulty was determined based on the number of objects and their size. 
CycleGAN was also trained to stylize source images to minimize style shift.

In this context, several UDA for object detection works seem to focus on feature alignment across domains.
Recent approaches mix up source and target images to conduct the adaptation procedure.
To the best of our knowledge, ConfMix~\cite{confmix} is the first approach in this direction.
ConfMix copy-pastes confident regions along with its detections from the target image to the source image, and supervises the model with a self-supervised loss on the mixed image, scaled by a confidence-induced factor, as well as a supervised loss the source image.
Although ConfMix is effective, we argue that this is a rather preliminary approach for UDA and there is significant room for improvement.
For example, the source image is used redundantly during training: the network processes it nearly twice at each iteration, once for the ``source pass", and once for the ``target pass" in its mix up form (see Fig.~\ref{fig:teaser}).
We believe that this may hinder the network to adapt its statistics to those of the target domain.
Moreover, ConfMix does not leverage data augmentation on the target domain to make adaptation more effective.
\ourmethod trains a detector with confident target image regions only. 
We believe that the source data is unnecessary and potentially deleterious during the target pass.

\subsection{Mixing under domain shift}

The benefit of data augmentation is evident in prior art \cite{aug1, aug2}. 
Cutout \cite{cutout} is a regularization strategy that models object occlusion. 
It simply masks out random square regions of training images and has shown to improve the overall performance of deep models. 
Mixup \cite{mixup} produces element-wise convex combinations of pairs of image examples and their labels. 
These techniques have been common to address regularization issues in vision tasks so far \cite{smmix}. 
Regarding augmentation-based UDA for object detection, the literature is still developing \cite{lossmix, parauda}. 
For instance, CutMix \cite{cutmix} was devised primarily for classification purposes, and then tailored to object detection \cite{confmix}. 
It replaces a random portion of an image with a randomly sampled portion from another image. 
However, this may override important source information (e.g.~objects and their bounding boxes, as well as background context) for the adaptation procedure.
Moreover, the region that is randomly cropped from the target may not carry detections to be used as pseudo-labels. 
ConfMix~\cite{confmix} partially addresses the above problem by selecting a confident region from the target instead of a random one (see Fig.~\ref{fig:teaser}).
See description of ConfMix and its limitations in the previous section.

\ourmethod aims to settle the shortcomings of mix up approaches at once. 
The target image combines random augmentations of a confident target region only.
We empirically show that our approach achieves state-of-the-art results.



%% file: sections/method.tex
\section{Our approach}\label{sec:approach} 

\input{images/pipeline/pipeline}

\subsection{Overview}

Given a detector $\Phi_\Theta$ pre-trained on source data, \ourmethod performs adaptation by self-training $\Phi_\Theta$ on challenging composite images obtained by combining different augmentations of the same crop extracted from target data.
We adapt the detector to the target distribution by imposing consistency between the detections predicted on the composite image and the detections of the most confident image region transformed accordingly to the augmentations.

\subsection{Detect, augment, compose, and adapt}

We define a pair of annotated source data as $(\textbf{X}_S, \textbf{G}_S)$, where $\textbf{X}_S \in \mathbb{R}^{H\times W \times3}$ is an RGB image sampled from the source data distribution $S$ and $\textbf{G}_S = \{ (\textbf{B}_i, c_i) : i = 1, \dots, N_S \}$ is the list of its $N_S$ ground-truth bounding boxes, where $\textbf{B}_i \in \mathbb{R}^4$ contains the coordinates of two opposite corners of the $i$th bounding box, and $c_i \in [ 1, \dots, C ]$ is a label representing the category of the object contained inside it.
We denote an RGB image sampled from the target data distribution $T$ as $\textbf{X}_T \in \mathbb{R}^{H\times W \times3}$.
Without loss of generality, we assume $\textbf{X}_S$ and $\textbf{X}_T$ to have the same size.
A neural network model designed to perform object detection can be defined as a parametric function $\Phi_\Theta \colon \mathbb{R}^{H \times W \times 3} \to \mathbb{R}^{M \times 4} \times [1, \dots, C]^{M}$ with learnable parameters $\Theta$ that takes as input an RGB image $\textbf{X}$ and produces as output a list of $M$ detections of the objects of interest contained in it, i.e. $\Phi_\Theta(\textbf{X}) = \textbf{D}$.

\subsubsection{Finding reliable pseudo-detections}
Given a target image $\textbf{X}_T$, we first feed it to the detector $\Phi_\Theta$ (pre-trained on source data), to obtain the detections $\textbf{P}_T \in \mathbb{R}^{M_T \times 4} \times [1, \dots, C]^{M_T}$.
Because we want to use $\textbf{P}_T$ as pseudo-labels to self-train our model, it is convenient to find a way to retain good detections and filter out the ones with low confidence which are more likely to be false positives.
To this end, we first divide the input image $\textbf{X}_T$ into $S_\text{row}$ rows and $S_\text{col}$ columns, creating a grid of $S_\text{row} \times S_\text{col}$ cells.
We then assign to each cell $\textbf{X}_T^{ij} \in \mathbb{R}^{ H/S_\text{row} \times W/S_\text{col} \times 3 }$, $i=1, \dots, S_\text{row}$, $j=1, \dots, S_\text{col}$, a confidence value obtained by averaging the confidence of the bounding boxes whose center belongs to it.
The most confident cell $\tilde{\textbf{X}}_T$ is cropped from $\textbf{X}_T$ and the portions of its bounding box detections that fall outside the selected region are trimmed accordingly to obtain $\tilde{\textbf{P}}_T$ (Fig.~\ref{fig:pipeline}, Crop).

\subsubsection{Composing augmented versions}
Since the most reliable detections $\tilde{\textbf{P}}_T$ occupy only a small portion of $\textbf{X}_T$, we perform the following operations: 
(i) we fill the remaining cells with useful information to avoid wasting computation, 
(ii) we sample such information from the target domain to balance source and target data during adaptation, and 
(iii) we fill remaining cells with augmented versions of $\tilde{\textbf{X}}_T$ and preserve pseudo-labels reliability because we transform $\tilde{\textbf{P}}_T$ according to the augmentation function chosen, as opposed to estimating them over potentially highly distorted versions of $\tilde{\textbf{X}}_T$.
To this end, we first perturb the pair $(\tilde{\textbf{X}}_T, \tilde{\textbf{P}}_T)$ by different data augmentations $f_1, f_2, \dots, f_{S_\text{row}\,S_\text{col}}$, randomly selected from a list of predefined ones (Fig.~\ref{fig:pipeline}, Augm.).
Throughout our experiments the list of predefined data augmentations includes horizontal flipping, cropping, blurring, color jittering, downscaling, and perturbing brightness and contrast (Tab.~\ref{tab:list_augm}), but it can be expanded or tailored according to the downstream task.
A composite image $\hat{\textbf{X}}_T = g \left( \tilde{\textbf{X}}_T \right) \in \mathbb{R}^{H \times W \times 3}$ is then created by composing the augmented versions of $\tilde{\textbf{X}}_T$ in a $S_\text{row} \times S_\text{col}$ grid as
\begin{equation*}\label{eq:comp}
    g \colon \mathbb{R}^{H/S_\text{row} \times W/S_\text{col} \times 3} \to \mathbb{R}^{H \times W \times 3},
    \quad
    g \left( \tilde{\textbf{X}}_T \right) = \begin{bmatrix}
    f_1 \left( \tilde{\textbf{X}}_T \right) & \dots & f_{S_\text{col}} \left( \tilde{\textbf{X}}_T \right) \\
    \vdots & \ddots & \vdots \\
    f_{(S_\text{row}-1) S_\text{col} +1} \left( \tilde{\textbf{X}}_T \right) & \dots & f_{S_\text{row} S_\text{col}} \left( \tilde{\textbf{X}}_T \right) \\
    \end{bmatrix},
\end{equation*}
obtaining an image of the same size of $\textbf{X}_T$ (Fig.~\ref{fig:pipeline}, Comp.).
Composite pseudo-detections are obtained in the same way, i.e. $\hat{\textbf{P}}_T = g(\tilde{\textbf{P}}_T)$ (Fig.~\ref{fig:pipeline}, Comp.).

\subsubsection{Loss function}
The core idea behind \ourmethod is to use $\hat{\textbf{P}}_T$ as pseudo-detections to self-train $\Phi_\Theta$.
To do so, we pass $\hat{\textbf{X}}_T$ in input to $\Phi_\Theta$ to obtain the detections $\hat{\textbf{D}}_T$ and promote consistency between $\hat{\textbf{D}}_T$ and $\hat{\textbf{P}}_T$ by minimizing the loss $\ell_T$.
Performing adaptation by minimizing $\ell_T$ alone may incur in two problems: (i) catastrophic forgetting \cite{catastrophicforgetting}, i.e. the knowledge learned by the detector on $S$ during the pre-training phase can be quickly forgotten, and (ii) potential false positive pseudo-detections on target data can undermine the performance.
To cope with both issues, during the adaptation phase we minimize the combined loss $\ell = \ell_S + \ell_T$, where $\ell_S$ promotes consistency between source detections $\textbf{D}_S = \Phi_\Theta(\textbf{X}_S)$ and its ground-truth annotations $\textbf{G}_S$ (Fig.~\ref{fig:pipeline}, first row).

%% file: images/pipeline/pipeline.tex
\begin{figure}[t!]
\centering

\begin{tikzpicture}

%
%
\node[anchor=south west,inner sep=0] at (0.0,0.0) {\includegraphics[width=22mm]{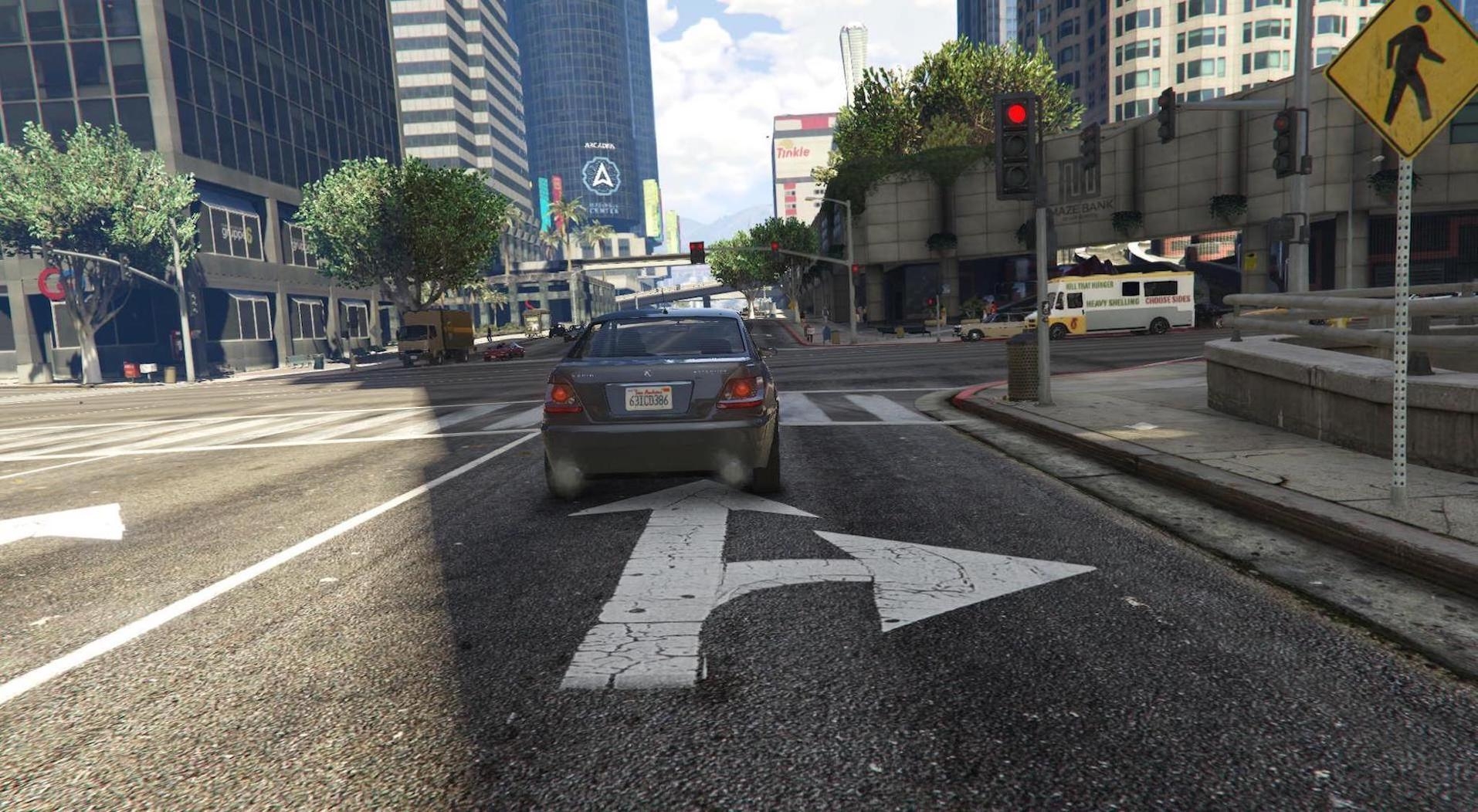}};
\node[shape=isosceles triangle, fill=blue!55, fill opacity=1.0, shape border uses incircle, shape border rotate=0, text height=9.4pt] at (2.67,0.61) {};
\node[shape=isosceles triangle, fill=blue!55, fill opacity=1.0, shape border uses incircle, shape border rotate=180, text height=9.4pt] at (3.32,0.61) {};
\node[inner xsep=1pt, inner ysep=3pt, outer sep=0pt, align=center, text opacity=1] at (3.0,0.61) {\small\color{white}{$\Phi_\Theta$}};
\node[anchor=south west,inner sep=0] (pred) at (3.78,0.0) {\includegraphics[width=22mm]{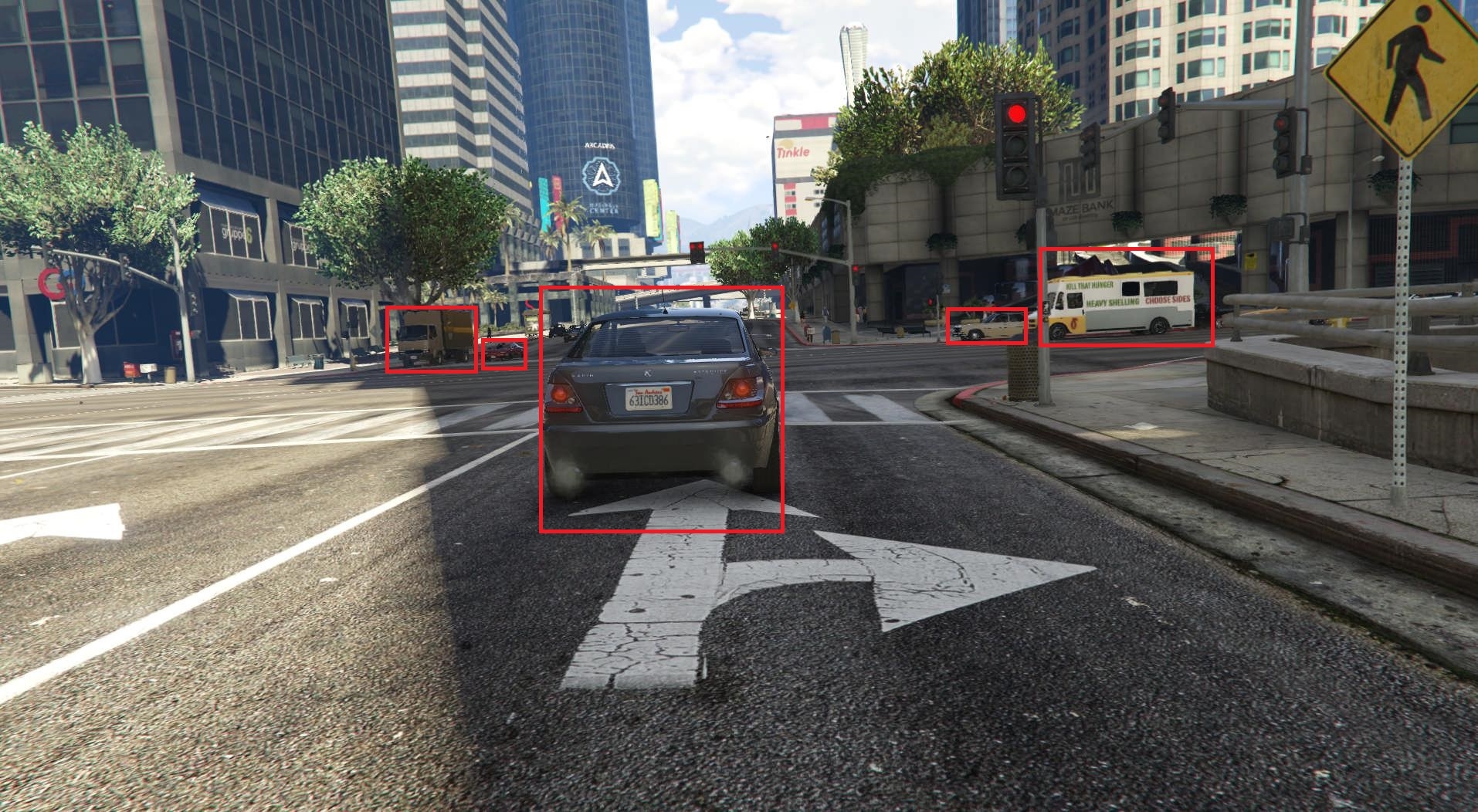}};
\node (loss) at (6.5,0.61) {\small $\ell_S$};
\node[anchor=south west,inner sep=0] (target) at (7.0,0.0) {\includegraphics[width=22mm]{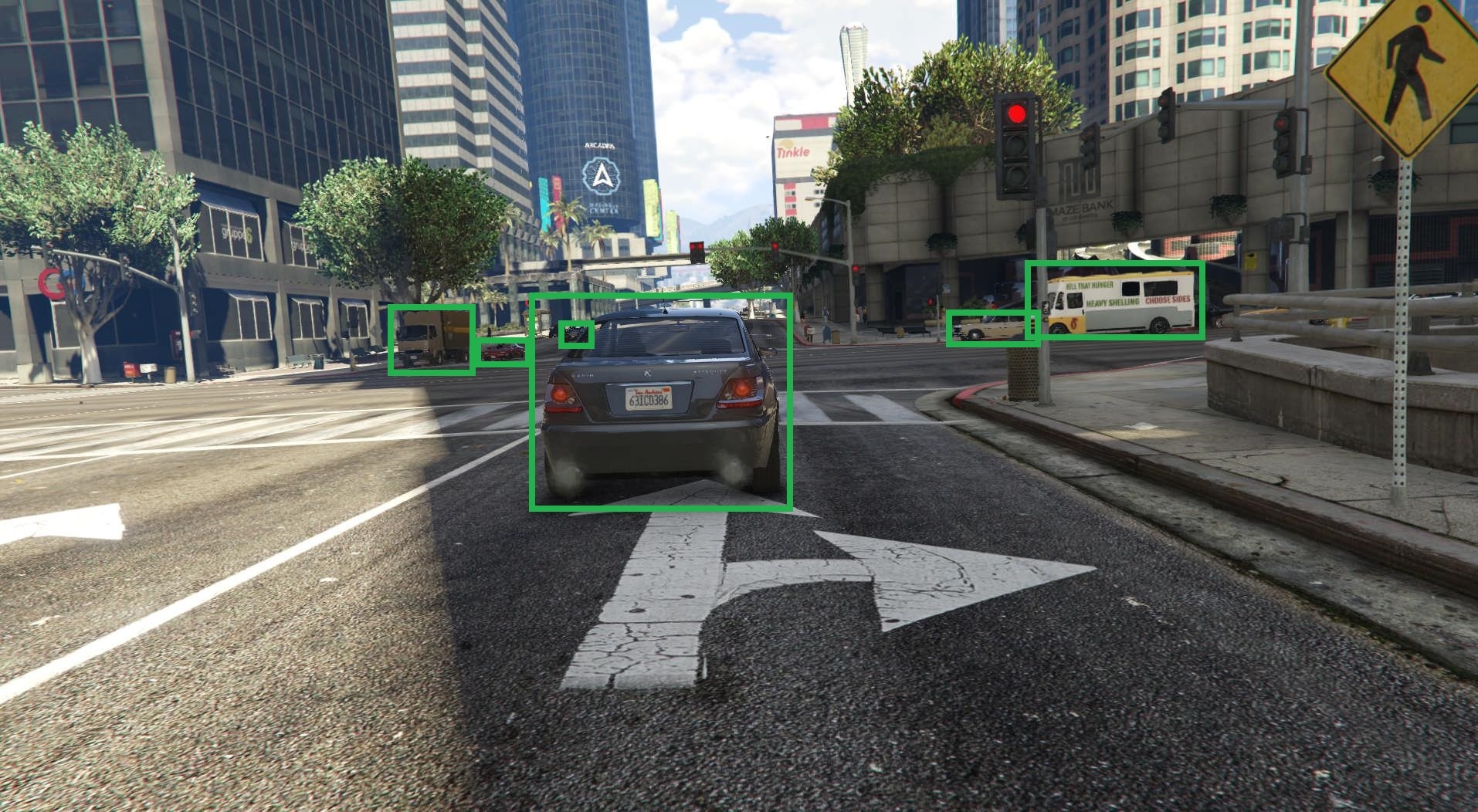}};
\draw[->] (pred) -- (loss);
\draw[->] (target) -- (loss);
\node at (1.1,-0.2) {\small $\textbf{X}_S$};
\node at (4.9,-0.2) {\small $\textbf{D}_S$};
\node at (8.1,-0.2) {\small $\textbf{G}_S$};

%
%
\node[anchor=south west,inner sep=0] at (0.0,-1.99) {\includegraphics[width=22mm]{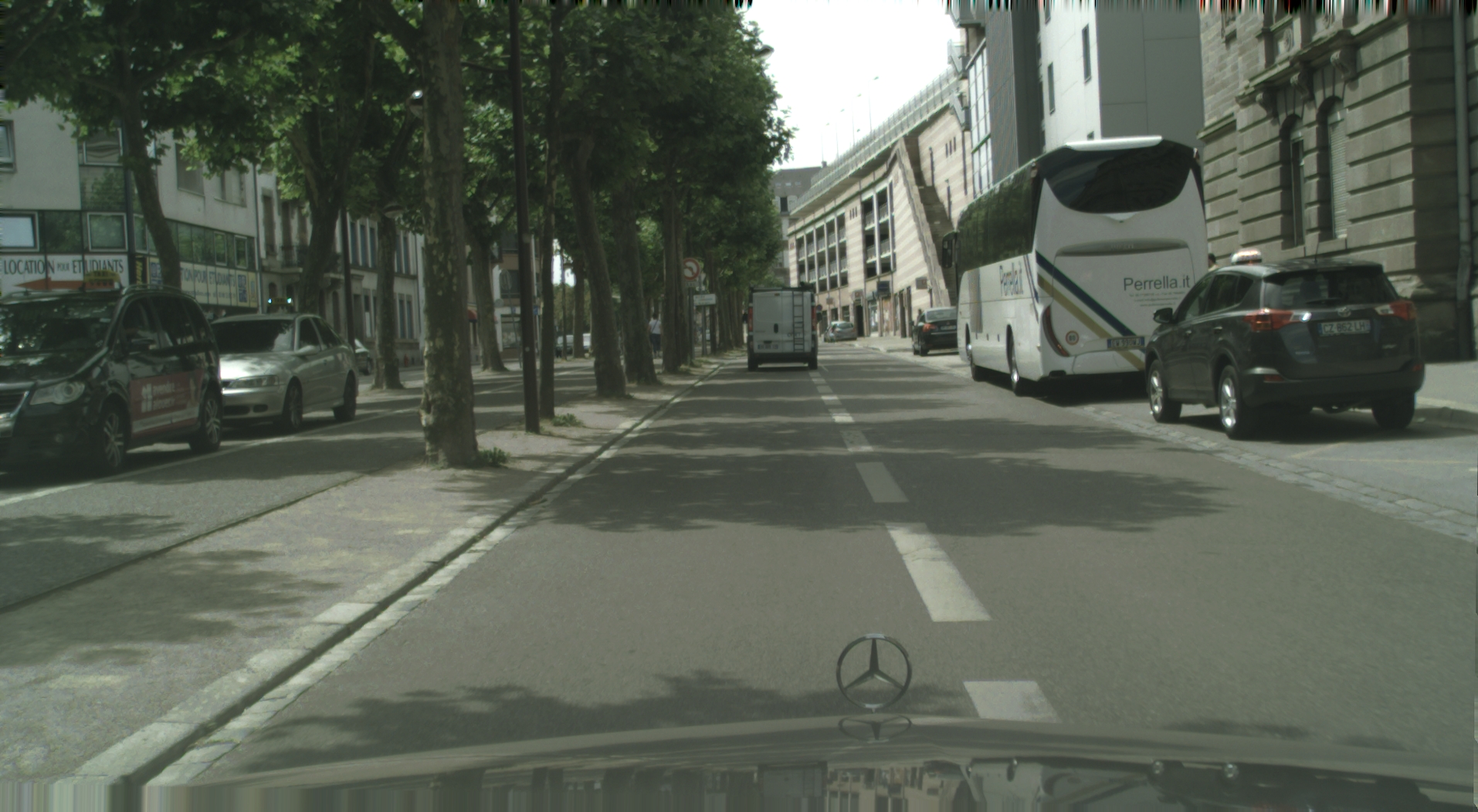}};
\node[shape=isosceles triangle, fill=blue!55, fill opacity=1.0, shape border uses incircle, shape border rotate=0, text height=9.4pt] at (2.67,-1.37) {};
\node[shape=isosceles triangle, fill=blue!55, fill opacity=1.0, shape border uses incircle, shape border rotate=180, text height=9.4pt] at (3.32,-1.37) {};
\node[inner xsep=1pt, inner ysep=3pt, outer sep=0pt, align=center, text opacity=1] at (3.0,-1.37) {\small\color{white}{$\Phi_\Theta$}};
\node[anchor=south west,inner sep=0] (pred) at (3.78,-1.99) {\includegraphics[width=22mm]{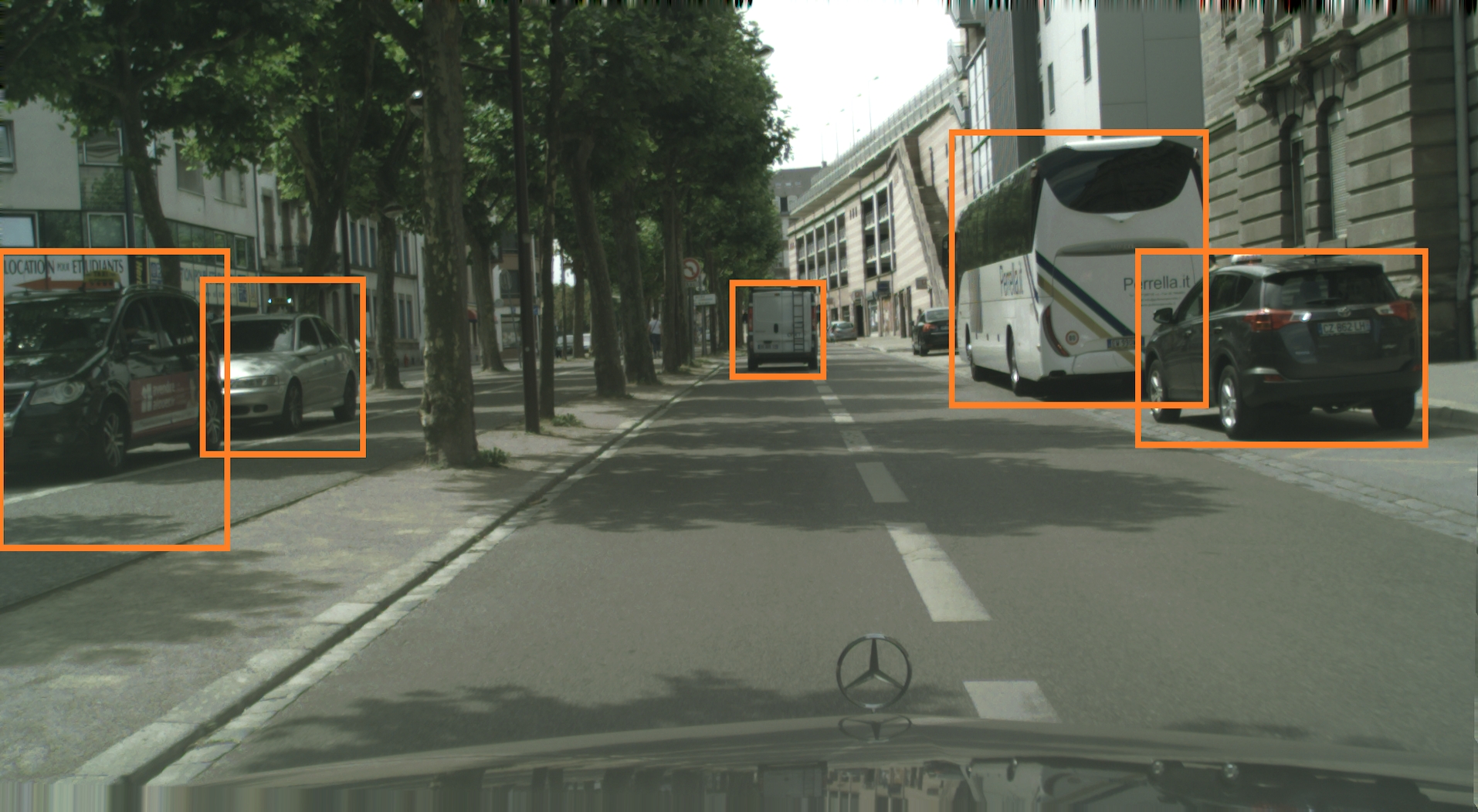}};
\node[inner xsep=1pt, inner ysep=8pt, outer sep=0pt, align=center, text opacity=1, fill=red!55, fill opacity=1.0] at (6.22,-1.38) {\small\color{white}\rotatebox{90}{Crop}};
\node[anchor=south west,inner sep=0, opacity=0.25] at (6.45,-1.99) {\includegraphics[width=22mm]{images/pipeline/trg_pseudo}};
\node[anchor=south west,inner sep=0] at (7.548,-1.385) {\includegraphics[width=11mm]{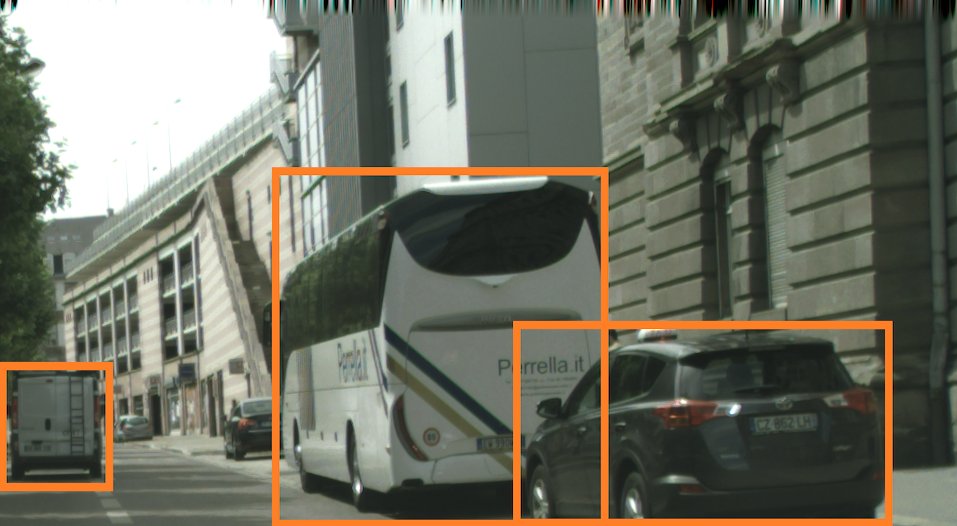}};
\draw[white, dashed, line width=0.75] (7.555,-1.99) -- (7.555,-0.77) {};
\draw[white, dashed, line width=0.75] (6.45,-1.375) -- (8.66,-1.375) {};
\node[inner xsep=1pt, inner ysep=4.75pt, outer sep=0pt, align=center, text opacity=1, fill=red!55, fill opacity=1.0] at (8.88,-1.382) {\small\color{white}\rotatebox{90}{Augm.}};
\node[anchor=south west,inner sep=0] at (9.1,-1.39) {\includegraphics[width=11mm]{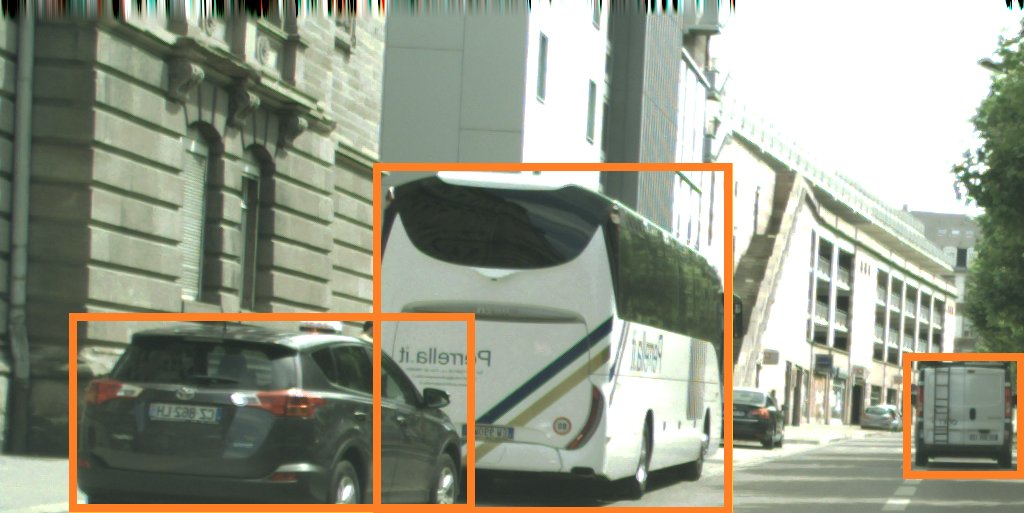}};
\node[anchor=south west,inner sep=0] at (10.3,-1.39) {\includegraphics[width=11mm]{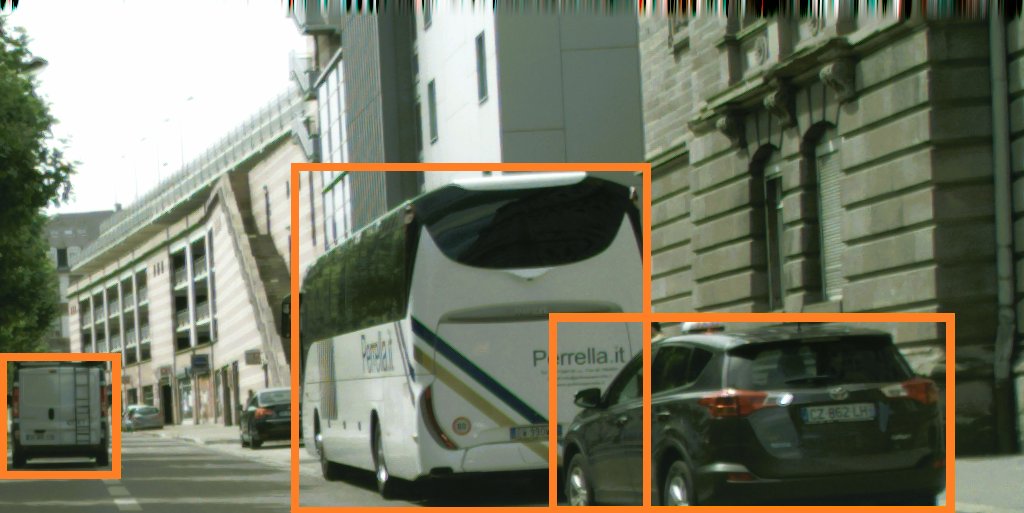}};
\node[anchor=south west,inner sep=0] at (9.7,-2.2) {\includegraphics[width=11mm]{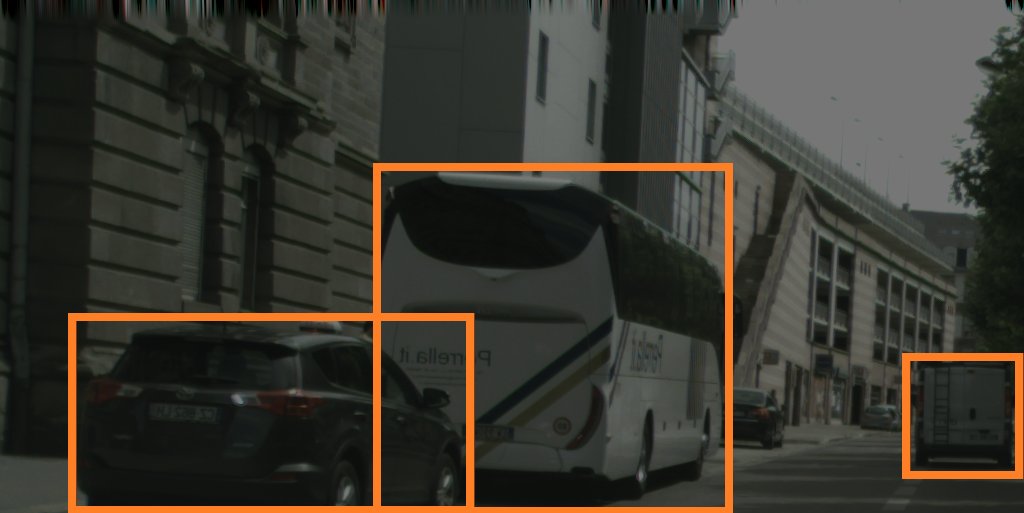}};
\node[anchor=south west,inner sep=0] at (10.9,-2.2) {\includegraphics[width=11mm]{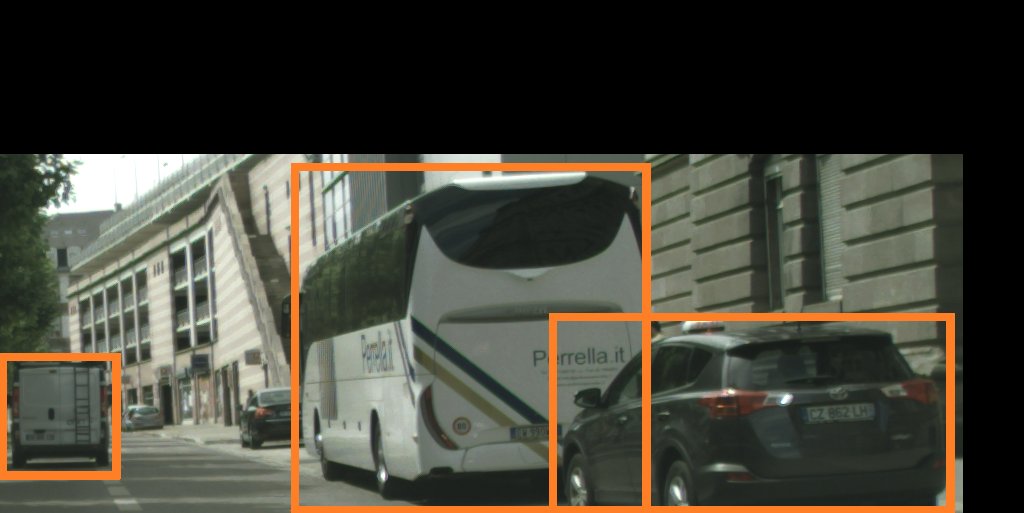}};
\node[anchor=south west,inner sep=0] at (10.2,-1.42) {\small ,};
\node[anchor=south west,inner sep=0] at (11.4,-1.42) {\small ,};
\node[anchor=south west,inner sep=0] at (10.8,-2.23) {\small ,};
\node[anchor=south west,inner sep=0] at (12.02,-2.2) {\small \dots};
\node[inner xsep=1pt, inner ysep=8pt, outer sep=0pt, align=center, text opacity=1, fill=red!55, fill opacity=1.0] (comp) at (12.6,-1.485) {\small\color{white}\rotatebox{90}{Comp.}};
\node at (1.1,-2.2) {\small $\textbf{X}_T$};
\node at (4.9,-2.2) {\small $\textbf{P}_T$};
\node at (8.15,-2.2) {\small $(\tilde{\textbf{X}}_T, \tilde{\textbf{P}}_T)$};

%
%
\node[anchor=south west,inner sep=0] (i2) at (0.0,-3.9) {\includegraphics[width=22mm]{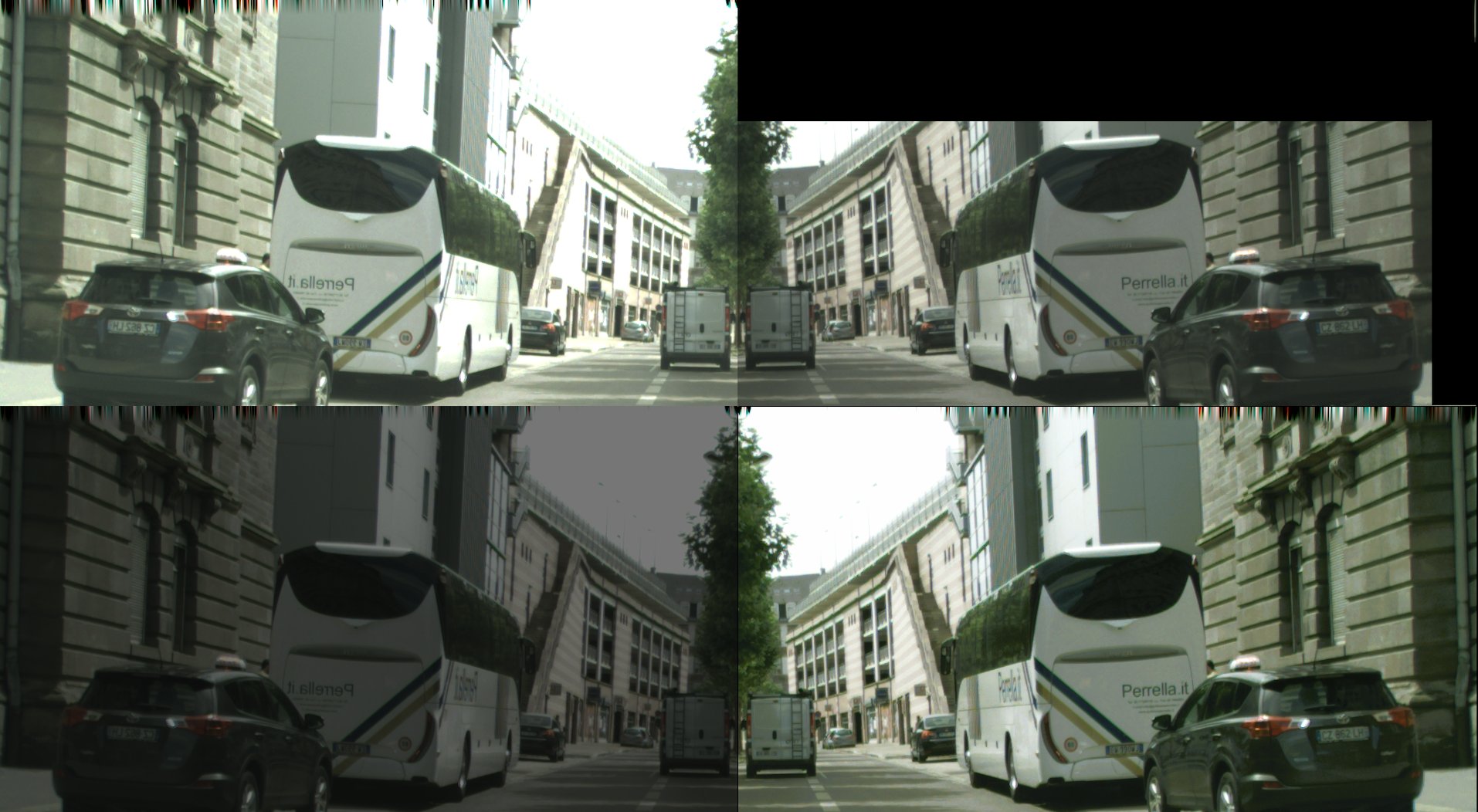}};
\node[shape=isosceles triangle, fill=blue!55, fill opacity=1.0, shape border uses incircle, shape border rotate=0, text height=9.4pt] at (2.67,-3.29) {};
\node[shape=isosceles triangle, fill=blue!55, fill opacity=1.0, shape border uses incircle, shape border rotate=180, text height=9.4pt] at (3.32,-3.29) {};
\node[inner xsep=1pt, inner ysep=3pt, outer sep=0pt, align=center, text opacity=1] at (3.0,-3.31) {\small\color{white}{$\Phi_\Theta$}};
\node[anchor=south west,inner sep=0] (p2) at (3.78,-3.9) {\includegraphics[width=22mm]{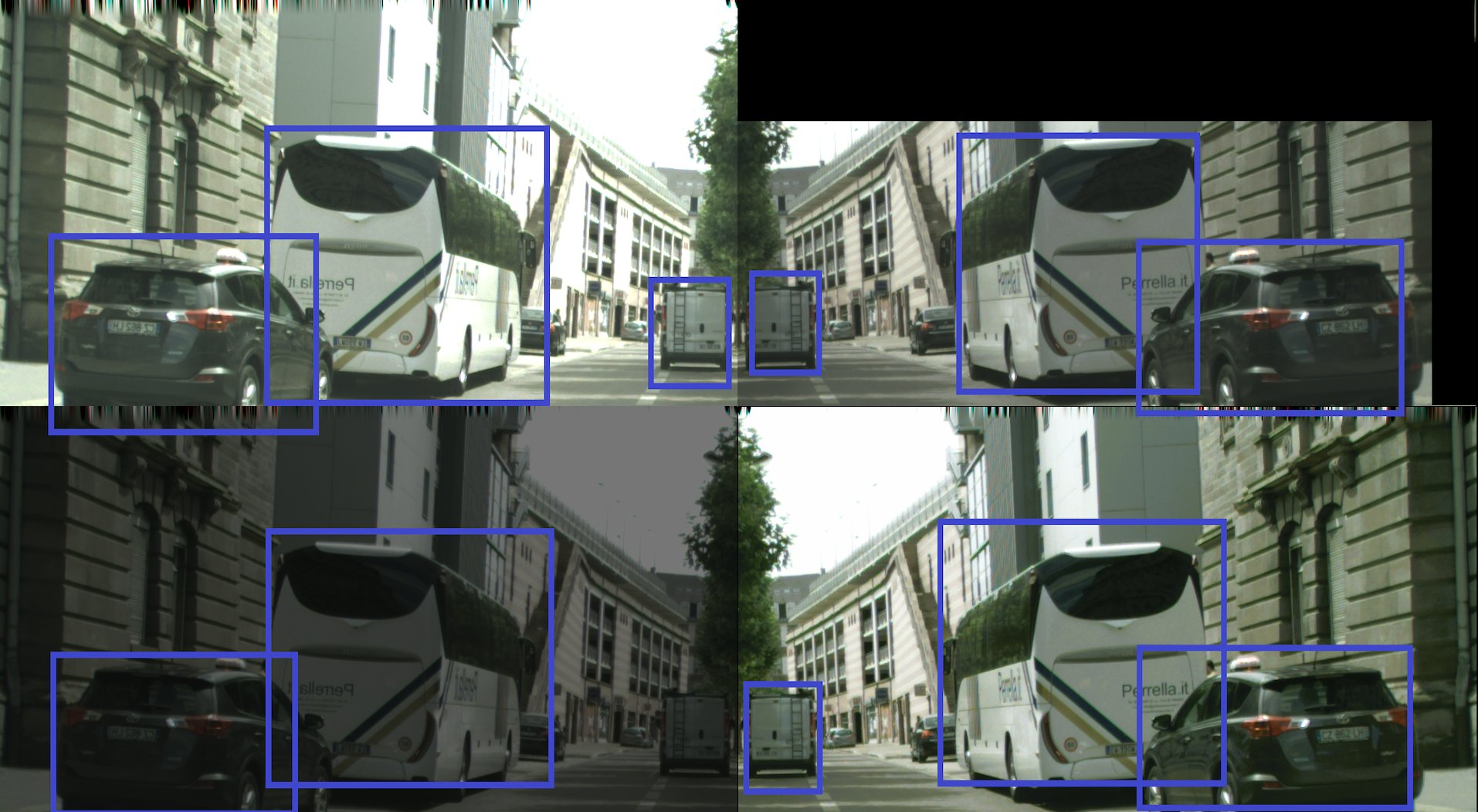}};
\node (l2) at (6.5,-3.31) {\small $\ell_T$};
\node[anchor=south west,inner sep=0] (t2) at (7.0,-3.9) {\includegraphics[width=22mm]{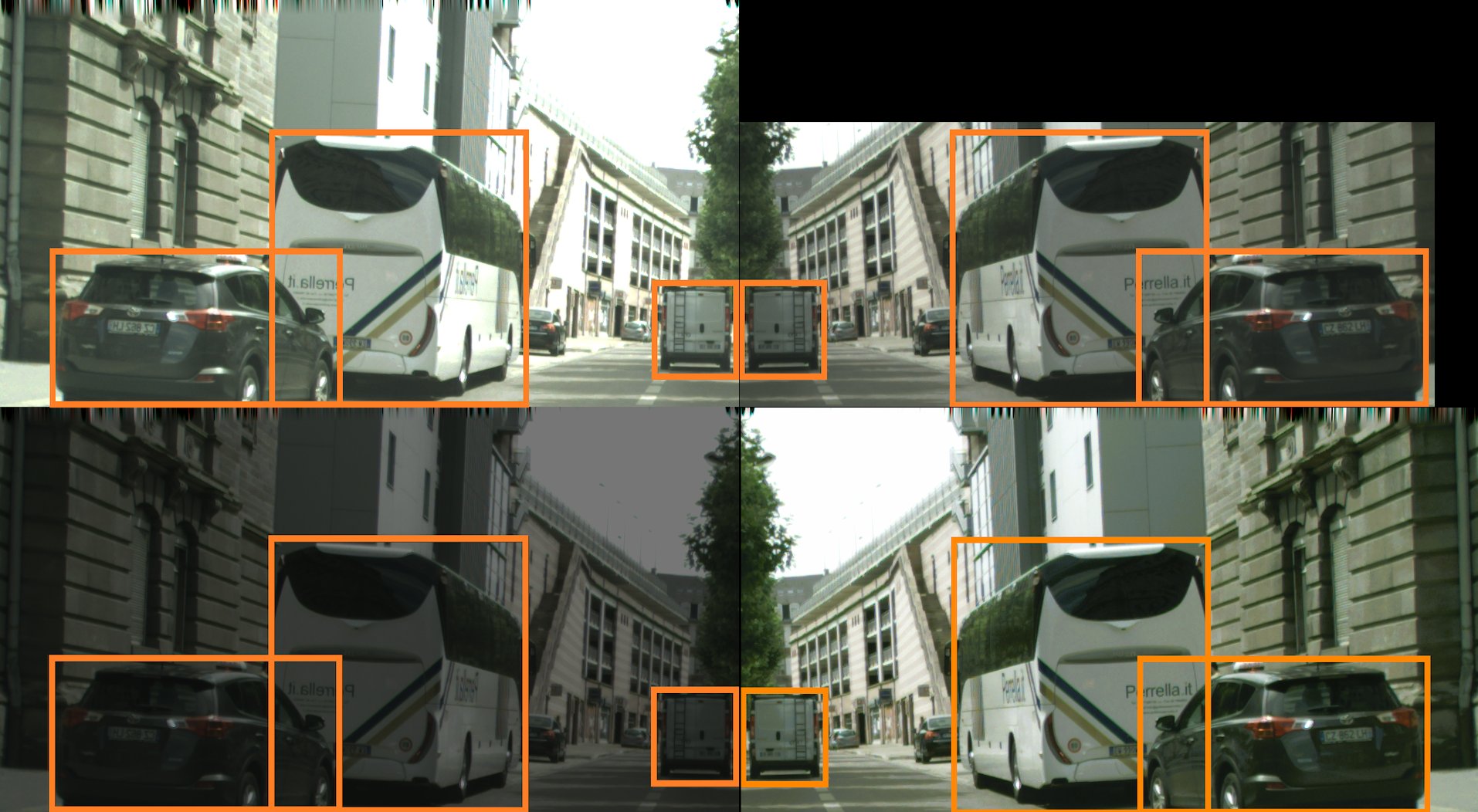}};
\draw[->] (comp.south) -- +(0mm,-2.5mm) -| (i2.north);
\draw[->] (comp.south) -- +(0mm,-2.5mm) -| (t2.north);
\draw[->] (p2) -- (l2);
\draw[->] (t2) -- (l2);
\node at (1.1,-4.1) {\small $\hat{\textbf{X}}_T$};
\node at (4.9,-4.1) {\small $\hat{\textbf{D}}_T$};
\node at (8.1,-4.1) {\small $\hat{\textbf{P}}_T$};

\end{tikzpicture}

\vspace{-2mm}
\caption{
Overview of our proposed approach (\ourmethod). 
The detector maintains its knowledge on the source by supervision via $\ell_S$, and jointly adapts towards the target via $\ell_T$ using the augmented target image and its pseudo-detections.
Keys: Comp.: composition. Augm.: augmentation.
Bounding boxes colors: red: source detections, green: source ground-truth, orange: target pseudo-detections, blue: composite target detections.
}
\label{fig:pipeline}

\end{figure}
\raggedbottom

%% file: sections/results.tex
\section{Experiments}\label{sec:results}

\subsection{Datasets}

We evaluate and compare our method under three benchmark adaptation scenarios, which include four datasets: 
(i) Cityscapes (C)~\cite{cityscapes}: it contains urban images acquired from different cities. Annotations are provided for the eight object classes Person, Rider, Car, Truck, Bus, Train, Motorcycle, and Bicycle; 
(ii) FoggyCityscapes (F)~\cite{foggycityscapes}: it is a variation of Cityscapes obtained by applying a synthetic fog filter with three different intensity levels; 
(iii) Sim10K (S)~\cite{sim10k}: it is a synthetic dataset produced by the GTA-V game engine;
(iv) KITTI (K)~\cite{kitti}: it is a traffic scenes dataset.
The first experimental scenario envisions a synthetic-to-real adaptation task on the Car category where $S$ is Sim10K and $T$ is Cityscapes (S$\to$C). 
The second experimental scenario is cross-camera adaptation on the Car category where $S$ is KITTI and $T$ is Cityscapes (K$\to$C). 
The last experimental scenario is weather-adaptation on Cityscapes' eight categories (C$\to$F).

\subsection{Implementation details}

We compare our method with previous works \cite{irgg, gipa, ssod, ctrp, cdn, fluda, SCUDA, sapn, megacda, mga, epm, scan, sigma} that used Faster R-CNN \cite{fasterrcnn} and FCOS \cite{fcos} detectors for their evaluation.
It is worth-mentioning that, for fair comparison in UDA, it is also important to compare based on the same detector, backbone and code reproduction details (e.g.~initial seeds). 
We directly compare with ConfMix~\cite{confmix}, as it is one of the most recent UDA methods for object detection.
In particular, we use the one-stage detector YOLOv5s as in \cite{confmix}, using COCO pre-trained weights as initialization and with a confidence threshold of 0.25 to select pseudo-labels.
We adopt a batch size of 2 including one source image and one target image, with an image size of 600$\times$600.
We train our detector with source data for 20 epochs, and then train with source and target data for 50 more epochs by using \ourmethod.
This number of epochs is the same as in \cite{confmix}.
We use Average Precision (AP) and Mean Average Precision (mAP) as evaluation metrics for all the experiments with the Intersection over Union threshold set to 0.5.

\input{tables/augm_types}

\subsection{Quantitative results}

We report the baseline scores (Source only) in which the model is trained on the source data and applied on the test set of the target, which represents our lower bound of the performance range. 
We also report the Oracle scores, where the model is trained on the target using its ground truth and tested on the test set of the target, which represents our upper bound of the performance range. 
We start the experiments with a 2$\times$2 target grid division layout in which the confident target region is augmented four times based on a list of typical image augmentations as listed in Tab.~\ref{tab:list_augm}.

\input{tables/quant_car}

Tab.~\ref{tab:quant_car} shows the quantitative results obtained on the S$\to$C, K$\to$C, and C$\to$F benchmarks for the Car category. 
\ourmethod outperforms all the previous works. 
We observe gains of $4.4\%$ and $2.6\%$ in the S$\to$C and K$\to$C cases, respectively, compared to the closest competitor method (ConfMix), achieving new state-of-the-art on both scenarios.
In particular, across the three scenarios, our method improves over ConfMix by $2.4\%$ on average.

Tab.~\ref{tab:quant_cf} shows the quantitative results obtained on the C$\to$F benchmark.
\ourmethod improves by $0.3\%$ over ConfMix, while both underperform some of other methods that use different detectors and different backbones (e.g.~MGA and SIGMA \cite{mga, sigma}).
Although this adaptation case is particularly challenging due to the high occlusion of objects and high domain shift owed to the presence of fog (confirmed also by the low mAP achieved by the baseline), the gap between \ourmethod and the oracle score is reasonably narrow ($3.3\%$).

\input{tables/quant_cf}

We display qualitative examples on the class Car for the K$\to$C and S$\to$S scenarios in Fig.~\ref{fig:qualitative_results} to highlight the fact that \ourmethod produces less false positives compared to ConfMix. 

More results can be found in the Supplementary Material.

\input{images/qualitative_results/qualitative_results}

\subsection{Ablation study}

\subsubsection{Effect of different augmentation types}
Tab.~\ref{tab:abl} (left) shows an ablation study on the six augmentation types listed in Tab.~\ref{tab:list_augm} and their combinations.
We can observe that horizontal flipping (HF) performs better than the other augmentations on average.
Downscale (D) and blurring (B) are the next best performing augmentations, they improve the model's robustness to scale and occlusion.
Combining all the augmentations together performs the best in all benchmarks.
Interestingly, turning off all the augmentations and using raw images (the None case in Tab.~\ref{tab:abl} (left)) lowers the mAP significantly, which underlines the benefit of using data augmentations for UDA.

\subsubsection{Effect of grid layout}
Tab.~\ref{tab:abl} (top right) shows an ablation on the grid layout used to find reliable detections for our best setting (the All case in Tab.~\ref{tab:abl} (left)).
We can see that a 3$\times$3 grid scores the lowest mAP, we believe this happens because it produces a small confident region and consequently few objects and small spatial context.
On the other hand, a 3$\times$2 grid is slightly better than a 2$\times$3 one, suggesting that the horizontal context is more informative than the vertical one due to the presence of cars and roads in the target datasets.
Finally, a 2$\times$2 grid performs best, being a good trade-off of the previous layouts.

\subsubsection{Does the number of augmented regions matter?}
Tab.~\ref{tab:abl} (bottom right) shows an ablation on the number of augmented regions for our best setting (the All case in Tab.~\ref{tab:abl} (left)).
We can observe that lowering the number of augmented regions that are used to build the composite image incurs a drastic performance decay. 
Fig.~\ref{fig:numregions} shows some qualitative results obtained in the S$\to$C adaptation scenario when using only one augmented region versus when four regions are used in a 2$\times$2 grid layout.
Using only one region compromises the robustness of the detector and yields more false positive detections.

\input{tables/abl}

\input{images/qualitative_results/numregions}

%% file: tables/augm_types.tex
\begin{table}[t]
\centering
\resizebox{1\columnwidth}{!}{
\begin{tabular}{llll}
\toprule
Transformation & Acronym & Operation & Hyper-parameters \\
\midrule
HorizontalFlip & HF & Randomly flips the image horizontally & probability($p$)=0.5 \\
BBoxSafeRandomCrop & SRC & Randomly crops the image without compromising & $p$=0.2 \\
& & the bounding boxes & \\
Blur & B & Randomly blurs out the image & $p$=0.5 \\
ColorJitter & CJ & Randomly applies color jitter & $p$=0.5, brightness=0.2, contrast=0.2, saturation=0.2, hue=0.2 \\
Downscale & D & Randomly downscales the image & $p$=0.5, minimum scale=0.5, maximum scale=0.99 \\
RandomBrightnessContrast & BC & Randomly alters brightness/contrast of the image & $p$=0.5, brightness limit=0.1, contrast limit=0.1 \\
\bottomrule
\end{tabular}
}
\vspace{-2mm}
\caption{List of image augmentation operations to build our composite input image.}
\label{tab:list_augm}
\end{table}
\raggedbottom

%% file: tables/quant_car.tex
\begin{table}[t]
\centering
\vspace{2mm}
\resizebox{.8\columnwidth}{!}{
\begin{tabular}{lllcccc}
\toprule
Method & Detector & Backbone & S$\to$C & K$\to$C & C$\to$F & Avg. \\
\midrule
CTRP \cite{ctrp} &
    Faster R-CNN & VGG-16 & 44.5 & 43.6 & 50.1 & 46.1\\
CDN \cite{cdn} &
    Faster R-CNN & VGG-16 & 49.3 & 44.9 & 50.9 & 48.4\\
FL-UDA \cite{fluda} &
    Faster R-CNN & VGG-16 & 43.1 & 44.6 & 44.4 & 44.0\\
SC-UDA \cite{SCUDA} &
    Faster R-CNN & VGG-16 & 52.4 & 46.4 & 56.0 & 51.6\\
SAPN \cite{sapn} &
    Faster R-CNN & VGG-16 & 44.9 & 43.4 & 59.8 & 49.4\\
MeGA-CDA \cite{megacda} &
    Faster R-CNN & VGG-16 & 44.8 & 43.0 & 52.4 & 46.7\\
MGA \cite{mga} &
    Faster R-CNN & VGG-16 & 54.6 & 48.5 & 60.6 & 54.6\\
IRGG \cite{irgg} &
    Faster R-CNN & ResNet-50 & 43.2 & 45.7 & 51.9 & 46.9\\
GIPA \cite{gipa} &
    Faster R-CNN & ResNet-50 & 47.6 & 47.9 & 54.1 & 49.9\\
SSOD \cite{ssod} &
    Faster R-CNN & ResNet-50 & 49.3 & 47.6 & 57.2 & 51.4\\
\midrule
EPM \cite{epm} &
    FCOS & ResNet-101 & 51.2 & 45.0 & 57.1 & 51.1\\
SCAN \cite{scan} &
    FCOS & VGG-16 & 52.6 & 45.8 & 57.3 & 51.9\\
SIGMA \cite{sigma} &
    FCOS & VGG-16 & 53.7 & 45.8 & 63.7 & 54.4\\
\midrule
Baseline (Source only) &
    YOLOv5 & CSP-Darknet53 & 50.4 & 42.9 & 54.9 & 49.4 \\
Oracle (Target only) &
    YOLOv5 & CSP-Darknet53 & 69.5 & 69.5 & 67.9 & 69.0 \\
ConfMix \cite{confmix} &
    YOLOv5 & CSP-Darknet53 & 56.2 & 51.6 & \textbf{63.0} & 56.9 \\
\ourmethod (Ours) &
    YOLOv5 & CSP-Darknet53 & \textbf{60.6} & \textbf{54.2} & \textbf{63.0} & \textbf{59.3} \\
\bottomrule
\end{tabular}
}
\vspace{2mm}
\caption{Comparison of Car detection performance (AP) on the S$\to$C, K$\to$C, and C$\to$F adaptation benchmarks. 
Keys:
C: Cityscapes. 
F: FoggyCityscapes. 
S: Sim10K. 
K: KITTI.
Avg.: average across the three adaptation scenarios.}
\label{tab:quant_car}
\end{table}
\raggedbottom

%% file: tables/quant_cf.tex
\begin{table}[hbt!]
\centering
\tabcolsep 3pt
\vspace{2mm}
\resizebox{.99\columnwidth}{!}{
\begin{tabular}{lcccccccccccc}
\toprule
Method & Detector & Backbone & Person & Rider & Car & Truck & Bus & Train & Motorcycle & Bicycle & mAP \\
\midrule
IRGG \cite{irgg} &
    Faster R-CNN & ResNet-50 & 37.4 & 45.2 & 51.9 & 24.4 & 39.6 & 25.2 & 31.5 & 41.6 & 37.1 \\
GIPA \cite{gipa} &
    Faster R-CNN & ResNet-50 & 32.9 & 46.7 & 54.1 & 24.7 & 45.7 & 41.1 & 32.4 & 38.7 & 39.5 \\
SSOD \cite{ssod} &
    Faster R-CNN & ResNet-50 & 38.8 & 45.9 & 57.2 & 29.9 & 50.2 & 51.9 & 31.9 & 40.9 & 43.3 \\
CTRP \cite{ctrp} &
    Faster R-CNN & VGG-16 & 32.7 & 44.4 & 50.1 & 21.7 & 45.6 & 25.4 & 30.1 & 36.8 & 35.9 \\
CDN \cite{cdn} &
    Faster R-CNN & VGG-16 & 35.8 & 45.7 & 50.9 & 30.1 & 42.5 & 29.8 & 30.8 & 36.5 & 36.6 \\
FL-UDA \cite{fluda} &
    Faster R-CNN & VGG-16 & 30.4 & 51.9 & 44.4 & 34.1 & 25.7 & 30.3 & 37.2 & 41.8 & 37.0 \\
SC-UDA \cite{SCUDA} &
    Faster R-CNN & VGG-16 & 38.5 & 43.7 & 56.0 & 27.1 & 43.8 & 29.7 & 31.2 & 39.5 & 38.7 \\
SAPN \cite{sapn} &
    Faster R-CNN & VGG-16 & 40.8 & 46.7 & 59.8 & 24.3 & 46.8 & 37.5 & 30.4 & 40.7 & 40.9 \\
MeGA-CDA \cite{megacda} &
    Faster R-CNN & VGG-16 & 37.7 & 49.0 & 52.4 & 25.4 & 49.2 & 46.9 & 34.5 & 39.0 & 41.8 \\
MGA \cite{mga} &
    Faster R-CNN & VGG-16 & 43.9 & 49.6 & 60.6 & 29.6 & 50.7 & 39.0 & 38.3 & 42.8 & 44.3 \\
\midrule
EPM \cite{epm} &
    FCOS & ResNet-101 & 41.5 & 43.6 & 57.1 & 29.4 & 44.9 & 39.7 & 29.0 & 36.1 & 40.2 \\
SCAN \cite{scan} &
    FCOS & VGG-16 & 41.7 & 43.9 & 57.3 & 28.7 & 48.6 & 48.7 & 31.0 & 37.3 & 42.1 \\
SIGMA \cite{sigma} &
    FCOS & ResNet-50 & 44.0 & 43.9 & 60.3 & 31.6 & 50.4 & 51.5 & 31.7 & 40.6 & 44.2 \\
\midrule
Baseline (Source only) &
    YOLOv5 & CSP-Darknet53 & 39.2 & 38.0 & 54.9 & 12.4 & 33.1 & 6.2 & 19.9 & 33.6 & 29.7 \\
Oracle (Target only) &
    YOLOv5 & CSP-Darknet53 & 45.6 & 43.0 & 67.9 & 30.2 & 48.0 & 39.4 & 30.3 & 37.5 & 42.7 \\
ConfMix \cite{confmix} &
    YOLOv5 & CSP-Darknet53 & 44.0 & 43.3 & 63.0 & 30.1 & 43.0 & 29.6 & 25.5 & 34.4 & 39.1 \\
\ourmethod (Ours) &
    YOLOv5 & CSP-Darknet53 & 41.9 & 40.8 & 63.0 & 29.4 & 42.2 & 37.2 & 27.8 & 33.0 & 39.4 \\
\bottomrule
\end{tabular}
}
\vspace{2mm}
\caption{Object detection performance (APs) on the C$\to$F scenario. 
The last column reports the mean average precision (mAP) across all the object categories. \ourmethod improves slightly over its direct competitor ConfMix \cite{confmix} while both score the highest AP on the Car class.}
\label{tab:quant_cf}
\end{table}
\raggedbottom

%% file: images/qualitative_results/qualitative_results.tex
\begin{figure}[t]
\centering
    \begin{tabular}{c c c}
        Groundtruth & 
        ConfMix~\cite{confmix} &
        \ourmethod (Ours) \\
        \rotatebox{90}{\tiny KITTI $\to$ Cityscapes}
        \includegraphics[width=0.3\textwidth]{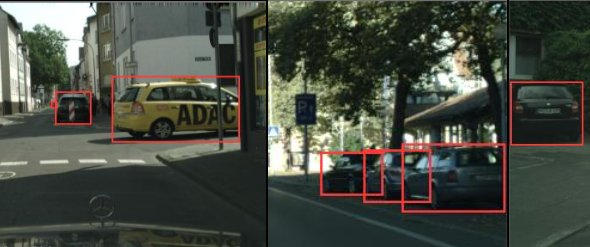} & \includegraphics[width=0.3\textwidth]{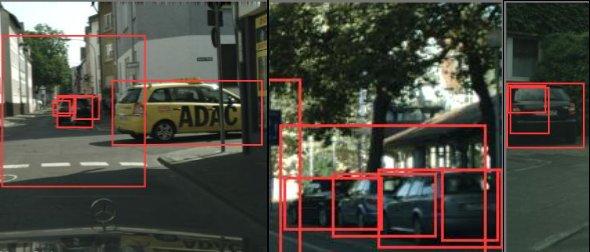} &  \includegraphics[width=0.3\textwidth]{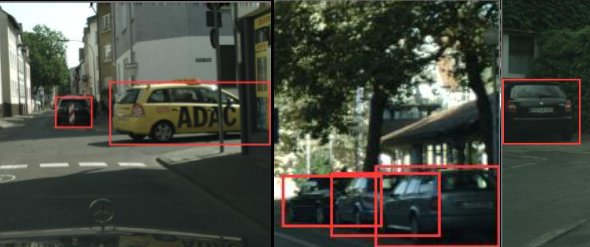}\\

        \rotatebox{90}{\tiny Sim10K $\to$ Cityscapes}
        \includegraphics[width=0.3\textwidth]{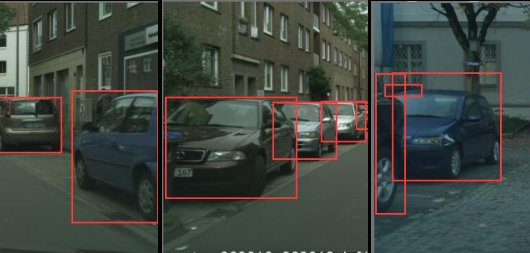} & \includegraphics[width=0.3\textwidth]{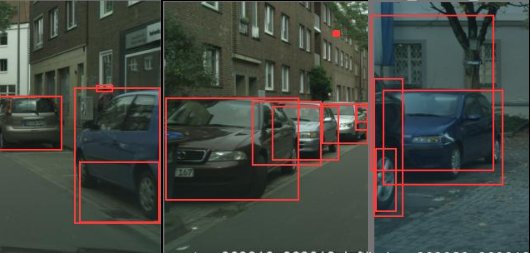} &  \includegraphics[width=0.3\textwidth]{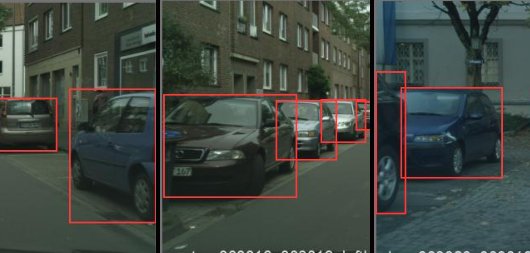}\\ 
       
    \end{tabular}
\vspace{-2mm}
\caption{
Qualitative instances of \ourmethod as compared to the ground truth and ConfMix \cite{confmix}.}
\label{fig:qualitative_results}
\end{figure}

%% file: tables/abl.tex
\begin{table}[t]
\centering
%
%
\begin{minipage}{0.45\textwidth}
\centering
\resizebox{.8\columnwidth}{!}{
\begin{tabular}{lcccc}
    \toprule
    Transf. & C$\to$F & K$\to$C & S$\to$C & Avg. \\
    \toprule
    None & 33.5 & 52.8 & 57.4 & 47.9 \\
    \midrule
    HF & 38.0 & 52.9 & 59.7 & 50.2 \\
    RC & 34.3 & 52.2 & 59.4 & 48.7 \\
    B & 35.9 & 53.2 & 58.4 & 49.2 \\
    CJ & 34.6 & 52.5 & 58.2 & 48.5 \\
    D & 35.3 & 54.1 & 59.8 & 49.8 \\
    BC & 33.9 & 52.6 & 57.5 & 48.0 \\
    \midrule
    HF+B & 38.9 & 53.9 & 59.3 & 50.7 \\
    HF+D & 35.4 & 53.3 & 56.7 & 48.5 \\
    D+B & 36.8 & 53.5 & 59.9 & 50.0 \\
    \midrule
    HF+D+B & 37.8 & 54.0 & 57.1 & 49.6 \\
    \midrule
    All & \textbf{39.4} & \textbf{54.2} & \textbf{60.6} & \textbf{51.4} \\
    \bottomrule
\end{tabular}
}
\end{minipage}
%
%
\begin{minipage}{0.45\textwidth}
\begin{minipage}[t!]{3.5cm}
\resizebox{1.4\columnwidth}{!}{
\begin{tabular}{ccccc}
    \toprule
    Grid layout & C$\to$F & K$\to$C & S$\to$C & Avg. \\
    \midrule
    3$\times$3 & 37.8 & 51.2 & 57.7 & 48.9 \\
    2$\times$3 & 38.5 & 51.7 & 58.7 & 49.6 \\
    3$\times$2 & 38.6 & 53.6 & 59.9 & 50.7 \\
    2$\times$2 & \textbf{39.4} & \textbf{54.2} & \textbf{60.6} & \textbf{51.4} \\
    \bottomrule
\end{tabular}
}
\end{minipage}

\vspace{7mm}
\begin{minipage}[b!]{3.5cm}
\resizebox{1.4\columnwidth}{!}{
\begin{tabular}{ccccc}
    \toprule
    Num. regions & C$\to$F & K$\to$C & S$\to$C & Avg. \\
    \midrule
    1 & 35.4 & 51.5 & 56.6 & 47.8 \\
    2 & 38.3 & 52.2 & 58.5 & 49.7 \\
    3 & 39.1 & 53.1 & 60.2 & 50.8 \\
    4 & \textbf{39.4} & \textbf{54.2} & \textbf{60.6} & \textbf{51.4} \\
    \bottomrule
\end{tabular}
}
\end{minipage}

\end{minipage}
\vspace{2mm}
\caption{Ablation results in terms of mAP for the three adaptation benchmarks.
Left: Effect of augmentation types.
Top right: Effect of grid layout.
Bottom right: Effect of the number of augmented regions (in a 2$\times$2 grid) to build the composite image.
C: Cityscapes, F: FoggyCityscapes, S: Sim10K, and K: KITTI.
Avg. Average across the three adaptation scenarios}
\label{tab:abl}
\end{table}
\raggedbottom

%% file: images/qualitative_results/numregions.tex
\begin{figure}[t]
\centering
    \begin{tabular}{ccc}
        Groundtruth & 
        One region &
        Four regions \\
        \rotatebox{90}{\tiny Sim10K $\to$ Cityscapes}
        \includegraphics[width=0.3\textwidth]{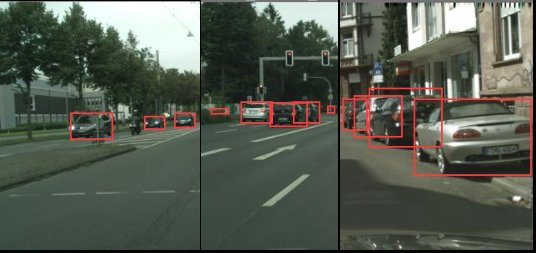} & \includegraphics[width=0.3\textwidth]{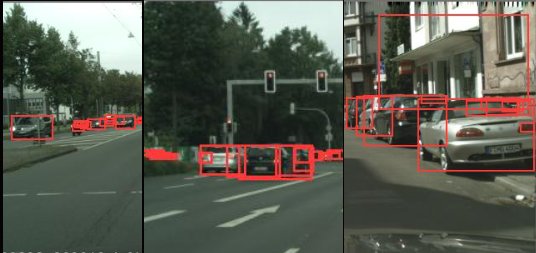} &  \includegraphics[width=0.3\textwidth]{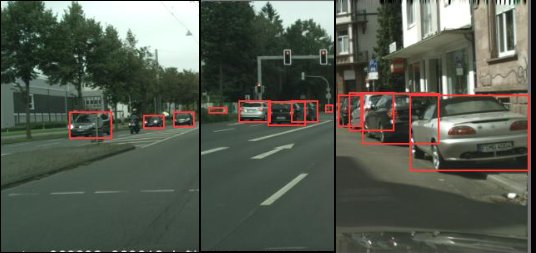}\\ 
    \end{tabular}
\vspace{-2mm}
\caption{Qualitative instances depicting the effect of the number of augmented regions used to build the composite image.}
\label{fig:numregions}
\end{figure}

%% file: sections/conclusions.tex
\section{Conclusions}

We presented a four-step approach for unsupervised domain adaptation in object detection.
We exploit regions of target images that have high-confidence object detections to collect pseudo-labels for self-training.
We generate a new image through the composition of different augmented versions of the selected image region and use its pseudo-labels to supervise the training of this composite image, which allows us to adapt the detector towards the target domain in a self-supervised fashion.
The adaptation of the detectors is carried out simultaneously with the training of source images to prevent knowledge drift of the detector.
We evaluated our approach on several benchmarks for UDA and compared it against alternative approaches.
Results show that our approach can outperform the other approaches.

As future work, one can explore style transfer techniques~\cite{fft} to improve \ourmethod's augmentation step, and consider confident bounding boxes instead of confident regions to improve \ourmethod's composition step by reducing the number of false positives negatively affecting the self-training process.


%% file: sections/acknow.tex
\section*{Acknowledgements}

We are very grateful to the support by European Union’s Horizon Europe research and innovation programme under grant agreement No. 101092043, project AGILEHAND (Smart Grading, Handling and Packaging Solutions for Soft and Deformable Products in Agile and Reconfigurable Lines).

%% file: sections/supp.tex
\newpage
\section*{Supplementary material}
Here, we provide more insights and carry out additional experiments in support of the main paper's content. In particular:
\begin{itemize}[noitemsep,topsep=0pt]
    \item We detail a step-by-step algorithm of how \ourmethod works.
    
    
    \item We assess the impact of the pseudo-label selection confidence threshold.

    \item We analyze the effect of different initialization schemes of the detector's weights. 
\end{itemize}

\setcounter{section}{0}
\setcounter{figure}{0}
\setcounter{table}{0}
\section{Algorithm}
Algorithm~\ref{alg:loop} details the key steps of \ourmethod.

\input{tables/sup_algo}

\section{Additional ablations}


\subsection{Pseudo-label selection}
Pseudo-label selection is a key step in \ourmethod. Intuitively, increasing the detector's confidence threshold imposes a strict pseudo-label selection criteria where only highly confident labels are selected and subsequently exploited for the adaptation routine. Conversely, decreasing the detector's confidence implies a loose selection mechanism where less confident pseudo-labels are leveraged. Therefore, it is necessary to evidence if high selection confidence is a better option than a low confidence.

For fairness of comparison, in the default experiments we kept the selection confidence set to 0.25 similarly to ConfMix \cite{confmix}. 
In this experiment, we further evaluate \ourmethod by using 0.1, 0.5 and 0.8 as confidences.
We also include Precision and Recall metrics to investigate the trend of False Positives and True Positives, respectively \cite{padilla}. 

From Tab.~\ref{tab:sup1}, we can notice that that using a selection threshold of 0.25 scores the highest mAP on average (51.4\%).
Decreasing the confidence threshold to 0.1 causes a 2.7\% decline, which is owed to the sharp decline in terms of Precision ($-$7.4\%).
This is a reasonable behavior provided that a low threshold entails more detections, which results in a higher rate of False Positives while favoring also True positive detections (i.e.~$+$1.5\% in terms of Recall).
By contrast, increasing the confidence threshold from 0.25 to 0.5 incurs fewer detections, which is manifested in a 8.8\% drop in terms of Recall, but also favors fewer False Positives ($+$2\% precision). 
When the selection threshold is further increased to 0.8, by far fewer detections are obtained (30.6\% Recall), which also affect significantly the Precision (59.2\%).
In conclusion, a threshold of 0.25 seems to be the best trade-off.
Figs.~\ref{fig:qual k2c}, \ref{fig:qual s2c}, and \ref{fig:qual c2f} depict comparison instances between the four selection threshold options from the three adaptation benchmarks. In particular, it can be observed that for a 0.1 confidence threshold, more False Positive detections are obtained. However, when the confidence threshold is increased to 0.5, the False Positives are reduced but often at the cost of missed True positives.

\input{tables/sup_conf}

\input{images/sup/qual_k2c}
\input{images/sup/qual_s2c}
\input{images/sup/qual_c2f}

\subsection{Weight initialization}
\ourmethod leverages self-training to perform UDA.
Prior to adaptation, the detector first infers pseudo-labels that would serve as self-supervision to learn the target modality.
Yet, he quality of the pseudo-labels also depends on the initialization of the detector's weights.
As in \cite{confmix}, in the default experiments we train the detector on the source dataset for 20 epochs starting with the COCO pre-trained model, and then we perform adaptation for 50 more epochs.

In this experiment we compare three other initialization alternatives:
(i) We discard the pre-training phase on the source and perform adaptation from scratch (i.e.~no COCO pre-trained model);
(ii) We train the detector on the source for 20 epochs starting from scratch, and then do adaptation as in the default setting. 
We also report the baseline (i.e.~the detector is trained on the source dataset from scratch using its ground truth) and oracle (i.e.~the detector is trained on the target dataset from scratch using its ground truth) performances in this setting;
(iii)  We perform adaptation starting with COCO pre-trained detector weights.

As shown Tab.~\ref{tab:sup2}, training the model from scratch on the source and then adapting it entails a significant mAP decay compared to training the model on the source starting with COCO pre-trained weights. 
In particular, a 12.4\% decline is observed on average over the three adaptation benchmarks. 
Both the lower and upper bound (i.e.~baseline and oracle) scores have dropped drastically when training from scratch.

When adapting the model from scratch without pre-training on the source dataset, we get the lowest mAP of 28.6\% on average over the three adaptation scenarios. 
However, when adapting the model starting with COCO pre-trained weights, we obtain a mAP of 49\% on average, which is significantly higher, indicating again that pre-training plays a pivotal role in UDA. 

Another worth-mentioning fact is that performing adaptation without pre-training on the source dataset starting with COCO weights outperforms, by far (+10\% mAP), the case when the model is first trained on the source from scratch and then adapted. 
This is rationale since COCO dataset is larger and carries much more semantic information, which qualifies it better for transfer learning tasks. 
The default case of training on the source dataset starting with COCO weights and then adapting is 2.4\% higher in terms of mAP across the three adaptation scenarios and 5.9\% higher on the C$\to$F scenario, which explains that source data is also necessary when addressing UDA.

\input{tables/sup_init_model}

%% file: tables/sup_algo.tex
\renewcommand{\algorithmicrequire}{\textbf{Input:}}
\renewcommand{\algorithmicensure}{\textbf{Output:}}

\begin{algorithm}[H]
\caption{Pseudocode of \ourmethod.}
\label{alg:loop}
\begin{algorithmic}[1]

\Require{
$\Phi_\Theta$: detector; 
$(\textbf{X}_S, \textbf{G}_S)$: source image and its ground truth; 
$\textbf{X}_T$: target image; 
$g$: image transformation function;
$\ell$: loss function; 
$S_\text{row}$, $S_\text{col}$: number of rows and columns for grid division, respectively;
$N_\text{it}$: number of adaptation iterations.} 

\Ensure{$\Phi_\Theta$ adapted via \ourmethod.}

\For{$i = 1, \dots , N_\text{it}$}:     

\State{Compute $\textbf{D}_S \leftarrow \Phi_\Theta(\textbf{X}_S)$.}

\State{Compute $\ell_S \leftarrow \ell(\textbf{G}_S, \textbf{D}_S)$.}

\State{Compute $\textbf{P}_T \leftarrow \Phi_\Theta(\textbf{X}_T)$.}

\State{Divide $\textbf{X}_T$ into $S_\text{row} \times S_\text{col}$ regions.}

\State{Associate each detection in $\textbf{P}_T$ to a region based on the center of its bounding box.}

\State{Compute the average detection confidence of each region.}

\State{Select the region with the highest average confidence and use its detections as pseudo-labels $(\tilde{\textbf{X}}_T, \tilde{\textbf{P}}_T)$.}

\State{Generate $S_\text{row} \times S_\text{col}$ different (random) augmentations of the selected region and its pseudo-labels.}

\State{Merge the augmentation regions to obtain a composite image $\hat{\textbf{X}}_T$ along with its pseudo-labels $\hat{\textbf{P}}_T$.}

\State{Compute $\textbf{D}_T \leftarrow \Phi_\Theta(\hat{\textbf{X}}_T)$.}

\State {Compute $\ell_T \leftarrow \ell(\hat{\textbf{P}}_T, \textbf{D}_T)$.}

\State{Compute $\ell \leftarrow \ell_S + \ell_T$.}

\State {Minimize $\ell$ to find the optimal parameters $\Theta$ for $\Phi_\Theta$.}

\EndFor
\State \Return {Adapted $\Phi_\Theta$.}
\end{algorithmic}
\end{algorithm}

%% file: tables/sup_conf.tex

\begin{table}[t]
\begin{center}
\tabcolsep 4.5pt
\begin{tabular}{|c|ccc|ccc|ccc|ccc|}
    \cline{2-13}
     \multicolumn{1}{c|}{} &\multicolumn{3}{c|}{C $\to$ F} & \multicolumn{3}{c|}{K $\to$ C} & \multicolumn{3}{c|}{S $\to$ C} & \multicolumn{3}{c|}{Average}\\
    \hline
    Conf. & P & R & mAP & P & R & mAP & P & R & mAP & P & R & mAP\\
    \hline
0.1&	59.9&	36.7&	38.8&	71.5&	41.5&	50.6&	74.1& 52.4&	56.6&	68.5&	43.5&	48.7\\    
    \hline
0.25&	65.2&	35.6&	39.4&	81.0&	38.4&	54.2&	81.5&	52.0&	60.6& 75.9&	42.0&	51.4\\
    \hline
0.5&	66.4&	31.1&	36.6&	84.2&	33.6&	53.0&	83.2&	34.8&	49.6& 77.9&	33.2& 46.4\\
    \hline
0.8&	55.7&	21.4&	26.3&	43.2&	39.7&	47.6&	78.7&	30.8&	47.1& 59.2&	30.6&	40.3\\
    \hline
\end{tabular}
\end{center}
\vspace{-1mm}
\caption{Effect of pseudo-label selection confidence threshold on the performance for all the adaptation benchmarks. Keys. Conf: Confidence threshold for pseudo-label selection, P: Precision, R: Recall, mAP: Mean Average Precision, C: Cityscapes, F: FoggyCityscapes, S: Sim10K, and K: KITTI.}
\label{tab:sup1}
\end{table}

%% file: images/sup/qual_k2c.tex




\begin{figure}[ht]
   \begin{minipage}{0.48\textwidth}
     \centering
     \textbf{Groundtruth}\par\medskip
     \includegraphics[width=0.87\textwidth]{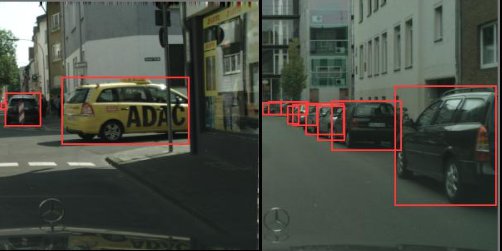}
   \end{minipage}
   \begin{minipage}{0.48\textwidth}
     \centering
     \textbf{\ourmethod}\par\medskip
     \rotatebox{90}{\tiny confidence = 0.25}
     \includegraphics[width=0.425\textwidth]
     {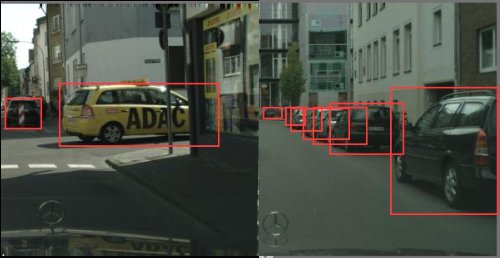} 
     \rotatebox{90}{\tiny confidence = 0.1}
     \includegraphics[width=0.435\textwidth]
     {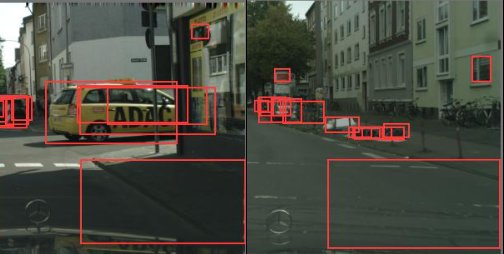}\\
     \rotatebox{90}{\tiny confidence = 0.5}
     \includegraphics[width=0.425\textwidth]
     {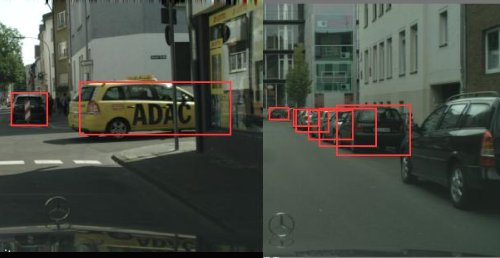} 
     \rotatebox{90}{\tiny confidence = 0.8}
     \includegraphics[width=0.43\textwidth]
     {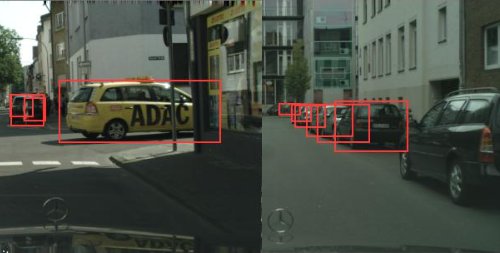}
   \end{minipage}
\vspace{2mm}
\caption{Qualitative instances of \ourmethod with different pseudo-label selection confidence thresholds for the K $\to$ C benchmark.}
\label{fig:qual k2c}
\end{figure}

%% file: images/sup/qual_s2c.tex




\begin{figure}[ht]
   \begin{minipage}{0.48\textwidth}
     \centering
     \textbf{Groundtruth}\par\medskip
     \includegraphics[width=0.87\textwidth]{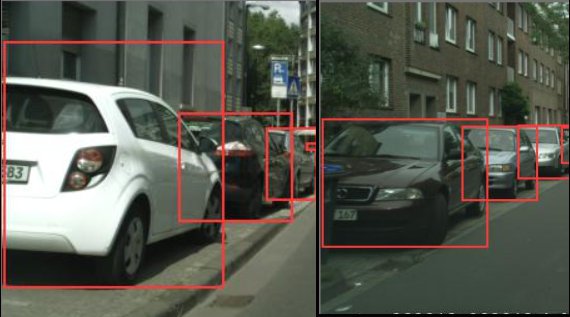}
   \end{minipage}
   \begin{minipage}{0.48\textwidth}
     \centering
     \textbf{\ourmethod}\par\medskip
     \rotatebox{90}{\tiny confidence = 0.25}
     \includegraphics[width=0.425\textwidth]
     {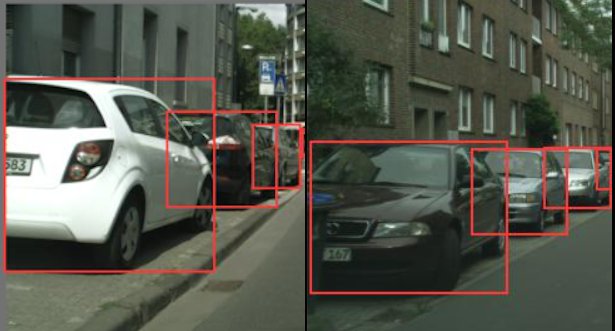} 
     \rotatebox{90}{\tiny confidence = 0.1}
     \includegraphics[height=0.237\textwidth]
     {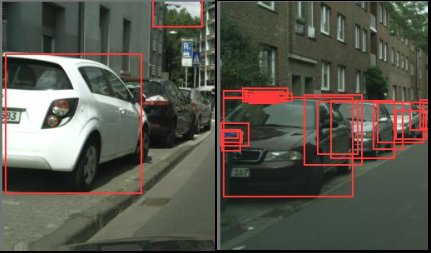}\\
     \rotatebox{90}{\tiny confidence = 0.5}
     \includegraphics[height=0.25\textwidth,
     width=0.425\textwidth]
     {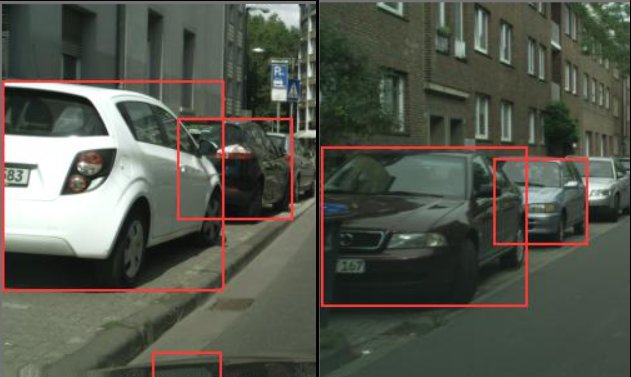} 
     \rotatebox{90}{\tiny confidence = 0.8}
     \includegraphics[height=0.25\textwidth, width=0.41\textwidth]
     {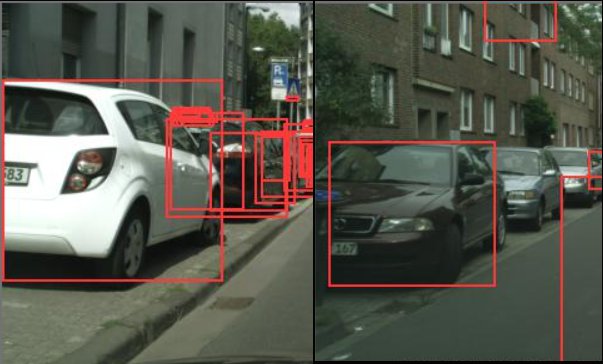}
   \end{minipage}
\vspace{2mm}
\caption{Qualitative instances of \ourmethod with different pseudo-label selection confidence thresholds for the S $\to$ C benchmark.}
\label{fig:qual s2c}
\end{figure}

%% file: images/sup/qual_c2f.tex




\begin{figure}[ht]
   \begin{minipage}{0.48\textwidth}
     \centering
     \textbf{Groundtruth}\par\medskip
     \includegraphics[width=0.87\textwidth]{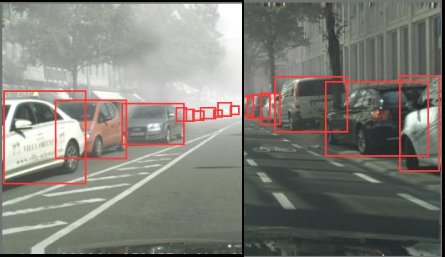}
   \end{minipage}
   \begin{minipage}{0.48\textwidth}
     \centering
     \textbf{\ourmethod}\par\medskip
     \rotatebox{90}{\tiny confidence = 0.25}
     \includegraphics[width=0.43\textwidth]
     {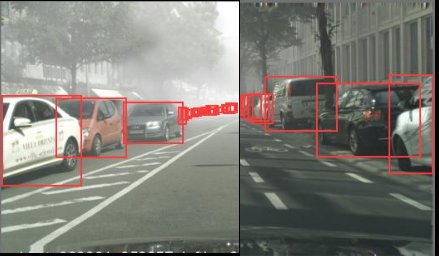} 
     \rotatebox{90}{\tiny confidence = 0.1}
     \includegraphics[width=0.43\textwidth]
     {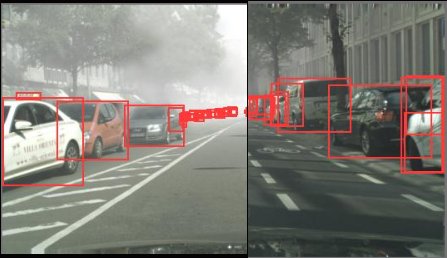}\\
     \rotatebox{90}{\tiny confidence = 0.5}
     \includegraphics[width=0.43\textwidth]
     {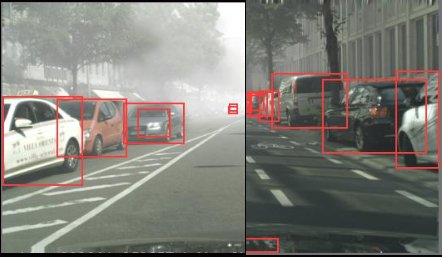} 
     \rotatebox{90}{\tiny confidence = 0.8}
     \includegraphics[width=0.43\textwidth]
     {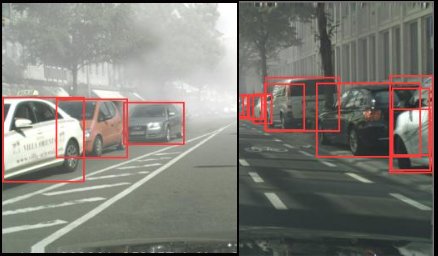}
   \end{minipage}
\vspace{2mm}
\caption{Qualitative instances of \ourmethod with different pseudo-label selection confidence thresholds for the C $\to$ F benchmark.}
\label{fig:qual c2f}
\end{figure}

%% file: tables/sup_init_model.tex
\begin{table}[H]
\centering
\tabcolsep 3pt
\resizebox{0.8\columnwidth}{!}{
\begin{tabular}{lccccc}
\toprule
Configuration & Pre-training & C$\to$F & K$\to$C & S$\to$C & Avg. \\
\toprule
%
%
Baseline (Source only) & None & 7.9 & 27.5 & 31.5 & 22.3 \\
\ourmethod (Adaptation on source and target) & None & 17.6 & 36.4 & 31.9 & 28.6 \\
\ourmethod (Finetuning on source + Adaptation on source and target) & None & 22.8 & 50.0 & 44.3 & 39.0 \\
Oracle (Target only) & None & 24.8 & 57.7 & 57.7 & 46.7 \\
\midrule
%
%
Baseline (Source only) & COCO & 29.7 & 42.9 & 50.4 & 41.0 \\
\ourmethod (Adaptation on source and target) & COCO & 33.5 & 53.7 & 60.0 & 49.0 \\
\ourmethod (Finetuning on source + Adaptation on source and target) & COCO & 39.4 & 54.2 & 60.6 &	51.4 \\
Oracle (Target only) & COCO & 42.7 & 69.5 & 69.5 & 60.6 \\
\bottomrule
\end{tabular}
}
\vspace{2mm}
\caption{Effect of different weights initialization schemes on the detection performance (mAP) for the three adaptation benchmarks. 
Keys:
C: Cityscapes,
F: FoggyCityscapes,
S: Sim10K,
K: KITTI,
Avg.: average across the three adaptation scenarios.}
\label{tab:sup2}
\end{table}